\pdfoutput=1
\documentclass{article}
\usepackage[T1]{fontenc}
\usepackage{iclr2027_conference,times}
\usepackage{amsmath, amssymb, booktabs}
\renewcommand{\toprule}[1][]{}
\usepackage{array,colortbl}
\definecolor{cost}{RGB}{42,91,123}
\definecolor{missing}{RGB}{145,150,155}
\definecolor{ours}{RGB}{235,244,249}
\newif\ifrevmark \revmarkfalse
\definecolor{revision}{RGB}{190,75,0}
\newcommand{\rev}[1]{\ifrevmark{\color{revision}#1}\else#1\fi}
\makeatletter
\newif\ifrev@outer
\newcommand{\rev@switch}[1]{%
  \ifnum\currentgrouplevel=\z@
    \ifrev@outer\reset@color\fi
    \global\rev@outertrue
  \fi
  #1}
\DeclareRobustCommand{\revbegin}{\ifrevmark\rev@switch{\color{revision}}\fi}
\DeclareRobustCommand{\revend}{\ifrevmark\rev@switch{\normalcolor}\fi}
\makeatother
\newcolumntype{L}[1]{>{\raggedright\arraybackslash}m{#1}}
\newcolumntype{C}[1]{>{\centering\arraybackslash}m{#1}}
\newcommand{\papertimingref}{Appendix~\ref{app:cost}}
\usepackage{graphicx}
\usepackage{placeins}
\usepackage[group-separator={,}]{siunitx}
\usepackage{tikz}
\usepackage[hidelinks]{hyperref}
\usepackage{url}
\title{Reinforcement Learning for Large-Scale\\
Lifelong Multi-Agent Path Finding\\
with Shared Spatial Fields}
\author{Jean Seong Bjorn Choe \\
Independent Researcher \\
\texttt{rnilva849@gmail.com}}
\date{}
\iclrfinalcopy
\begin{document}
\maketitle
\lhead{Preprint. Under review.}
\begin{abstract}
Large-scale applications of lifelong multi-agent path finding (LMAPF) call for computationally
inexpensive movement decisions under changing traffic conditions and across varied layouts.
We propose a navigation policy in which a convolutional network computes a shared spatial field
over the map once per timestep and a compact goal-conditioned operator derives agents' movement
preferences. The complete policy is trained end-to-end using average-reward reinforcement
learning with throughput as the reward, without planner demonstrations or a separate online
path planner. For structured and unstructured layouts, we train one policy per class on \revbegin{}pools\revend{}
of small maps and deploy each without retraining on larger, unseen layouts and agent populations.
Across six layouts with \num{10000} agents, our policies achieve throughput comparable to or
higher than the strongest evaluated baselines. The proposed method \rev{has an average execution time
below 10~ms per step}, more than an order of magnitude faster than several established planners.
\end{abstract}

\section{Introduction}
\label{sec:intro}

Lifelong multi-agent path finding (LMAPF) coordinates collision-free movement of agents that receive
new goals after completing their current ones. Throughput is the average number of goals completed
per timestep~\citep{ma2017lifelong,li2021rhcr}. Applications include large robotic logistics systems
that continually transport goods and real-time strategy games in which many units move toward
changing objectives. Large populations increase the computation needed to update movement decisions,
while real-time applications limit the time available for each update~\citep{silver2005cooperative}.
Reuse across facilities or game maps also requires handling varied obstacle and passage
geometry~\citep{stern2019mapf}. We therefore seek policies that maintain throughput across layouts
and populations with low per-step computation cost.

Existing methods use different forms of coordination within this computation limit. RHCR solves
windowed planning problems~\citep{li2021rhcr}, while WPPL refines windowed plans within an available
time budget~\citep{jiang2024wppl}. PIBT resolves conflicts through local priority rules~\citep{okumura2022pibt};
Traffic Flow Optimisation adds congestion-aware guide paths~\citep{chen2024trafficflow}.
Learned policies can replace part of this online planning. SILLM combines per-agent neural
inference, spatial communication and global guidance to coordinate \num{10000} agents~\citep{jiang2025sillm}.
When agents' local observations overlap, per-agent neural inference can repeatedly process the
same map and traffic information.

We use a shared spatial field to reuse this computation across agents. A convolutional network
processes the map and current traffic once per timestep, producing scores for movement and waiting
at each location. Agents use these shared scores in a goal-conditioned route computation. Neural
computation is performed over the map, while each agent's route computation is limited to a fixed
number of \rev{steps ahead} (Fig.~\ref{fig:actor}).

\begin{figure}[t]
\centering
\includegraphics[width=\linewidth]{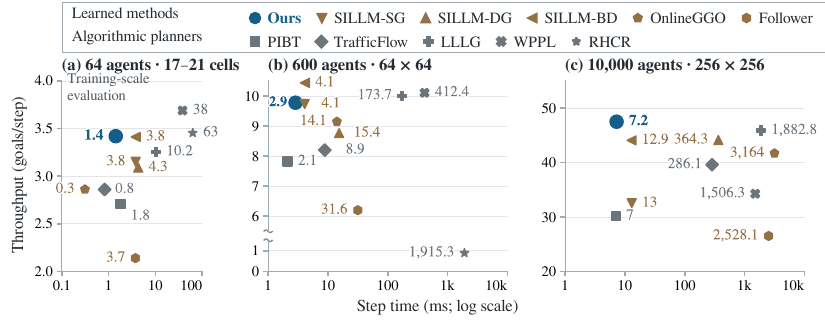}
\caption[Throughput and execution cost across random-map scales.]{\protect\input{figures/throughput_cost_compact_20260926/caption.tex}}
\label{fig:throughput-cost}
\end{figure}

A soft shortest-path distribution derived from the map and each agent's goal supplies a geometric
reference that favours shorter routes while retaining alternatives. The route operator combines
this reference with the shared field, accumulates scores along short route continuations, and
computes the resulting next-move probabilities. Parallel Sinkhorn iterations\rev{~\citep{sinkhorn1967concerning,cuturi2013sinkhorn}} then adjust these
probabilities for competing demand for destination cells. The spatial field, route operator and allocation define one navigation
policy.

We train this policy end-to-end using average-reward reinforcement learning. The population is
treated as one controlled system, with the total number of completed goals at each step as its
reward. Gradients through route reweighting and destination allocation train the convolutional
network. Training requires no planner demonstrations or additional task reward shaping
(Sec.~\ref{sec:training}).

We train one policy for structured layouts and one for unstructured layouts, each on a fixed \revbegin{}pool\revend{}
of small maps with roughly a hundred agents per map. The structured \revbegin{}pool\revend{} contains rack and
sortation layouts; the unstructured \revbegin{}pool\revend{} contains random maps, dumbbell layouts and street
crops. Each policy is applied without retraining to larger layouts excluded from training and to
different agent populations (Sec.~\ref{sec:experiments}).

Figure~\ref{fig:throughput-cost} summarises throughput and execution cost across random-map scales.
Across three training seeds per class at 25k updates, the unstructured policies reach
$95$--$99\%$ of RHCR's throughput on small random maps with $8$--$64$ agents and are reused
on $256\times256$ maps with $\num{2000}$--$\num{20000}$ agents. Across the six $\num{10000}$-agent
layouts, the policies attain throughput comparable to or higher than the highest-throughput
evaluated baseline on each layout, with execution below 10~ms per step and \revbegin{}$34$--$320\times$\revend{}
faster than that baseline (Sec.~\ref{sec:results}).

\section{Problem formulation}
\label{sec:prelim}

\paragraph{Continuing multi-agent control.}
We model LMAPF as a continuing Markov decision process on a four-connected grid \rev{$\mathcal G=(V,E)$} with
obstacles removed. For a fixed map and population of $N$ agents, the state $s_t$ contains agent
positions, \rev{each agent's current goal and the goal queued after it}, previous moves, elapsed-task counters and the task-stream state. The action is a joint movement
proposal $\mathbf a_t=(a_{1,t},\ldots,a_{N,t})$. At cell $x$, let $\mathcal M(x)$ be the moves to
adjacent free cells and $\mathcal A(x)=\mathcal M(x)\cup\{\mathrm{WAIT}\}$ the available actions;
$x^a$ denotes action $a$'s destination. Movement follows POGEMA's soft collision
rules~\citep{skrynnik2025pogema}, which prevent agents from occupying the same cell or exchanging
positions in one timestep.
Completing a goal assigns the next task, so operation continues~\citep{ma2017lifelong,li2021rhcr}.
These movement rules and task assignment define the transition law
$P(s_{t+1}\mid s_t,\mathbf a_t)$.

\paragraph{Throughput objective.}
The reward $r_t$ counts goals completed by all agents at step $t$. For a policy $\pi_\theta$, the
objective is long-run throughput for a fixed initial-state distribution,
\begin{equation}
 \rho(\theta)=\lim_{T\to\infty}\frac1T\,
 \mathbb E_{\pi_\theta}\!\left[\sum_{t=1}^{T}r_t\right].
 \label{eq:avgreward}
\end{equation}
The limit is assumed to exist. This continuing task has no terminal configuration in which all goals are complete. Average-reward
reinforcement learning targets the sustained completion rate; its value estimates describe future
reward relative to that rate~\citep{mahadevan1996average,sutton2018rl}.
Section~\ref{sec:training} specifies the estimator and treatment of training resets.

\paragraph{Soft shortest-path reference.}
Before considering traffic, map geometry and a goal $g$ define a reference action distribution.
Let $\widehat D_g(x)$ be a precomputed soft distance obtained by exponentially weighting the lengths
of walks to $g$. For cells with available movement actions,
\begin{equation}
 \begin{aligned}
 p^0(\mathrm{WAIT}\mid x,g)&=w_0,\qquad w_0=\frac{e^{-\beta}}{1+e^{-\beta}},\\
 p^0(a\mid x,g)&=(1-w_0)
 \frac{e^{-\beta\widehat D_g(x^a)}}
 {\sum_{b\in\mathcal M(x)}e^{-\beta\widehat D_g(x^b)}},\qquad a\in\mathcal M(x).
 \end{aligned}
 \label{eq:reference_action}
\end{equation}
We fix $\beta=3$. \rev{These probabilities are those of logit route choice in its Markov
form~\citep{akamatsu1996cyclic}, known in route choice modelling as the recursive logit
model~\citep{fosgerau2013rl}, when every step, a move or a wait, has the same utility. Under this model
the probability of a route to $g$ decreases exponentially with its number of steps.} The soft distances supply geometric route information beyond a single move,
while retaining alternatives to a shortest path. They contain no traffic information or learned
parameters. Appendix~\ref{app:reference} gives the distance construction and boundary rules.

\section{Approach}
\label{sec:actor}

\begin{figure}[t]
\centering
\includegraphics[width=\linewidth]{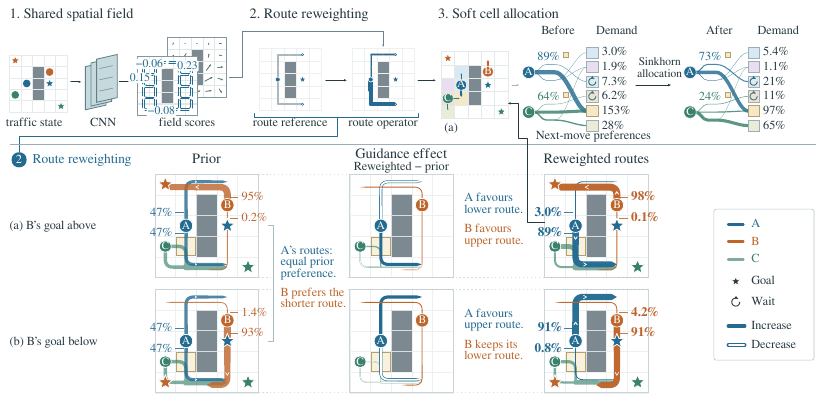}
\caption[Shared guidance, route reweighting and soft allocation.]{\protect\input{figures/figure1_combined_mechanism_v24/caption.tex}}
\label{fig:actor}
\end{figure}

The centralised policy computes a shared spatial field, goal-conditioned route probabilities and a destination
allocation (Fig.~\ref{fig:actor}). A convolutional network processes the scene once per timestep.
The route operator uses its output and each agent's goals to compute action probabilities, which
allocation adjusts for competing demand for cells.
\rev{For each agent the route operator considers routes of $H$ steps through the current scene,
with $H=4$ unless stated otherwise. We call one such step, a move or a
wait, a departure, and a sequence of $H$ departures a continuation. The depth
$h\in\{0,\ldots,H-1\}$ indexes the departures, with $h=0$ the next action.
All learned parameters are shared across agents and the network is convolutional, so the policy
applies unchanged to any population and map size.}
Appendix~\ref{app:illustration} specifies the example in Figure~\ref{fig:actor}.

\subsection{Shared spatial field}
\label{sec:medium}
\label{sec:params}

Let $O_t=O(s_t)$ be an aggregate spatial representation of the full state $s_t$, containing map geometry, agent positions and
previous moves, current and queued goal information, and \rev{counts of moves along routes sampled from the
agents' references}.
The convolutional network $W_{\theta_W}$ computes the shared field
$\Phi_t=W_{\theta_W}(O_t)$ once per timestep. \rev{This single evaluation serves all agents, so its
cost grows with the area of the map and not with the population.} \rev{At each cell the field holds scores for waiting and the four moves at every departure depth.
The route operator uses them as a bonus or penalty on each action at each cell, the same for all
agents, and to scale each agent's preference for its goal direction (Section~\ref{sec:read}).}
A base action-score vector is shared across depths, with additive corrections for later departures.
Scores enter the route probabilities exponentially, so higher accumulated scores favour the
corresponding continuations. All depths use the current scene; the field is refreshed at the next
timestep. \rev{Its main branch has two $3\times3$ convolutional layers of width 48 with multi-scale
output heads, and the whole network has 39,615 parameters. The route operator and
allocation add six further scalars.} Appendix~\ref{app:writer} specifies the observations, architecture and field construction.

\subsection{Goal-conditioned route operator}
\label{sec:read}

\paragraph{Goal-conditioned guidance.}
Agents with different goals can interpret the same field differently. A route state $z=(x,\varphi)$
records a candidate cell \rev{$x$ and a goal phase $\varphi$. A continuation that reaches the
agent's current goal continues toward its queued goal, and $\varphi$ records which of the two is
active.} Write $p_i^0(a\mid z)$ for
Eq.~\eqref{eq:reference_action} evaluated at agent $i$'s active goal.
Appendix~\ref{app:reference} specifies the goal-phase transitions and continuation boundary.
For action $a$, let $\psi_i(z,a)\in\mathbb R^2$ contain \rev{two preferences computed from the
reference probabilities of the movement actions, that for the direction of $a$ over the opposite
direction and that for the axis of $a$ over the perpendicular axis. For a move $a$ the two sum to its
reference probability minus the mean over the valid moves, and both are zero for waiting.}
The local guidance score is
\begin{equation}
 U_{i,h}(z,a)=b_h(x,a)+\gamma_h(x)^\top\psi_i(z,a).
 \label{eq:read}
\end{equation}
\rev{The field supplies both terms. Here $b_h(x,a)$ is its score for action $a$ at cell $x$ and
depth $h$, the same for every agent. The two gains $\gamma_h(x)$ are a linear function of the field's
waiting score and summed movement scores at $x$, with four learned coefficients shared across
agents, cells and depths, so that they need no separate network output (Eq.~\eqref{eq:routegain}).} The reference determines the direction of the adjustment; the field determines its
strength and sign. At an unobstructed junction, the \rev{preference for east over west} is
$\tfrac12[p_i^0(E\mid z)-p_i^0(W\mid z)]$, so a positive gain reinforces the direction favoured by
the reference in the local score. Appendix~\ref{app:reader} gives \rev{these preferences and the} field-to-gain map,
including obstacle masks.

\paragraph{Route reweighting.}
\label{sec:marginal}
For a continuation $\xi=(a_0,\ldots,a_{H-1})$ with successive route states $z_h$, the reference law
and cumulative guidance are
\begin{equation}
 P_i^0(\xi\mid z_0)=\prod_{h=0}^{H-1}p_i^0(a_h\mid z_h),\qquad
 S_i(\xi)=\sum_{h=0}^{H-1}U_{i,h}(z_h,a_h).
 \label{eq:reference_routes}
\end{equation}
The operator reweights these routes at learned temperature $\tau_f>0$ and returns their first-action marginal,
\begin{align}
 Q_i(\xi)&=\frac{P_i^0(\xi)\exp(S_i(\xi)/\tau_f)}
 {\sum_{\xi'}P_i^0(\xi')\exp(S_i(\xi')/\tau_f)},\label{eq:measure}\\
 q_i(a)&=\sum_{\xi:a_0=a}Q_i(\xi).\label{eq:logits}
\end{align}
\rev{In the terms of route choice, the scores $U_{i,h}/\tau_f$ are utilities added to the first $H$
departures of a route. The later steps keep the utilities of the reference, so the rest of the route to
the goal enters through the reference alone. Equivalently, $Q_i$ maximises cumulative guidance under a
KL penalty for departing from the reference,}
\begin{equation}
 Q_i=\arg\max_Q\left\{\mathbb E_Q[S_i(\xi)]-\tau_f\,\mathrm{KL}(Q\Vert P_i^0)\right\},
 \label{eq:route_choice}
\end{equation}
\rev{as in randomized shortest paths~\citep{saerens2009rsp,kivimaki2014rsp} and linearly solvable
control~\citep{todorov2006linearly}.} The optimisation ranges over distributions on the reference support. Throughput remains the objective
used to train the policy.

A backward log-sum-exp recursion computes $q_i$ over $H$ departures. Fixed $H$ bounds the per-agent
continuation computation independently of map size (Appendix~\ref{app:route_law}). Gradients of the next-move probabilities
can therefore update field scores along these continuations, beyond the agent's current cell.
\rev{The soft distances behind the reference are computed once per map and do not depend on traffic.
Current traffic enters the route probabilities only through the field, so no path or distance is recomputed
during execution.}
Section~\ref{sec:mechanism} compares policies trained with different continuation depths.

\subsection{Destination allocation}
\label{sec:projection}

The route operator supplies next-move probabilities $q_i$. To account for competing demand for
cells, we use affinities $A_i(a)=\tau_f\log q_i(a)$ in an entropy-regularized assignment between
agents and their one-step destinations~\citep{cuturi2013sinkhorn}. Each agent receives unit mass and
each cell has unit capacity. Additional rows represent unoccupied capacity; entropy applies to
both these rows and the agent assignments, at learned temperature $\tau_p>0$. Each agent connects
only to its valid one-step destinations, and the identical extra rows are represented collectively
without constructing a dense assignment matrix.

We apply Sinkhorn iterations twice, using a fixed iteration budget in each pass.
The first uses $A_i(a)$ to estimate a penalty
$c_i(a)$ for a likely move by the destination's occupant into agent $i$'s current cell. The second
uses $A_i(a)-c_i(a)$ and computes destination potentials $\lambda_x$ jointly across agents.
Row normalisation gives the proposal probabilities
\begin{equation}
 \pi_{\theta,i}(a\mid s_t)\propto q_i(a)^{\tau_f/\tau_p}
 \exp\!\left[-\frac{c_i(a)+\lambda_{x_i^a}}{\tau_p}\right].
 \label{eq:prices}
\end{equation}
Allocation also raises $q_i$ to the power $\tau_f/\tau_p$ before normalisation, independently of the
reverse-move penalties and
destination potentials. The potentials adjust demand for each cell. We abbreviate $\pi_{\theta,i}(a\mid s_t)$ as $\pi_i(a)$ below.

Training samples movement proposals from $\pi_i$; deployment chooses each agent's most probable
action. \rev{At a finite iteration count the cell capacities hold only approximately, and agents choose
independently, so proposals can still conflict. The collision rules of
Section~\ref{sec:prelim} cancel such moves.} Gradients pass through the route operator and final allocation pass,
with the first pass's reverse-move probabilities held fixed. Appendix~\ref{app:projection} gives
the allocation formulation and numerical procedure.

\subsection{Learning from throughput}
\label{sec:training}

We train the policy with AR-EAPO~\citep{choe2025areapo}, an average-reward actor--critic
method using a PPO surrogate~\citep{schulman2017ppo} and a separate entropy advantage. We train with the total goal
completion count defined in Section~\ref{sec:prelim} as the task reward, without planner demonstrations
or additional task reward shaping.

\paragraph{Average-reward estimation.}
For training-map entry $m$, $\widehat\rho_m^r$ estimates its reward rate and
$V_t^r=V^r_{\chi,m}(O_t^V)$ is the critic's differential value estimate from its
scene representation $O_t^V$. Let $V_{t+1}^{r,+}$ denote its value at the successor
before any reset. The reward TD residual is
\begin{equation}
 \delta_t^r=r_t-\widehat\rho_m^r+V_{t+1}^{r,+}-V_t^r.
 \label{eq:differential_td}
\end{equation}
Generalised advantage estimation~\citep{schulman2016gae} accumulates these residuals without temporal
discounting. A separate average-reward estimator uses the negative log probability of sampled proposals,
averaged over active agents, to estimate an entropy advantage. The reward and entropy advantages
are scaled separately and combined with entropy temperature $\tau$.
Per-map rate estimates and value-target normalisation accommodate different
throughput levels across the training \revbegin{}pool.\revend{} The critic is discarded at deployment.

\paragraph{Joint policy update.}
Conditional on the scene, agents sample their proposals independently from the probabilities
computed by the shared network, route operator and joint allocation. The log probability of the joint proposal is
\begin{equation}
 \log\pi_\theta(\mathbf a_t\mid s_t)=\sum_{i=1}^N\log\pi_{\theta,i}(a_{i,t}\mid s_t).
 \label{eq:joint_policy}
\end{equation}
The PPO update uses the corresponding joint likelihood ratio and one advantage for the full population, with an
adaptive KL penalty. Gradients through these proposal probabilities train the differentiable policy.

\paragraph{Training pools and continuing rollouts.}
\revbegin{}\label{sec:horizon}
We\revend{} train three seeds per fixed \revbegin{}pool,\revend{} using $H=4$ for the full policies.
The structured \revbegin{}pool\revend{} contains rack and sortation layouts, with uniform goals and alternating
workstation--storage tasks. The unstructured \revbegin{}pool\revend{} contains random maps, dumbbell layouts and
street crops from Denver, Moscow, Boston and Milan; Berlin and Paris are excluded.
States persist across updates. Independent random resets introduce fresh placements on the same
map and change the sampled state distribution. At a reset, value targets bootstrap from the ordinary
successor state before replacement and the advantage trace stops. Appendix~\ref{app:training} gives the full optimisation settings, reset rates,
pool composition and final weight-averaging procedure.

\section{Experiments}
\label{sec:experiments}
\label{sec:protocol}
\label{sec:selection}

We evaluate throughput and per-step execution cost on unseen layouts, then examine
transfer as agent populations change. We use the structured and unstructured policies from Section~\ref{sec:training},
with the terminal averaged checkpoint from each of three training seeds.
Unless stated otherwise, results average cases within each seed and then weight the seed means equally.
The six-layout benchmark uses \num{10000} agents, occupying 17--26\% of free
cells. Instances, task streams and \revbegin{}the number of timesteps\revend{} follow SILLM's released
evaluation protocol for every method. Appendix~\revbegin{}\ref{app:eval_protocol} describes the
maps and lists these settings.\revend{}

\paragraph{Baselines.}
SILLM~\citep{jiang2025sillm} combines learned action preferences and PIBT with
shortest-distance heuristics (BD), fixed directional edge costs (SG), or
TrafficFlow's congestion-aware guide paths (DG). OnlineGGO~\citep{zang2025onlineggo}
instead learns traffic-dependent edge costs for PIBT.
\rev{We use publicly released checkpoints for SILLM and Follower~\citep{skrynnik2024follower},
and train OnlineGGO once per configuration following its released recipe.
For our policies, we report variability across three independent training seeds
per class.} Table~\ref{tab:learned_baselines} summarises training and reuse.
Non-learned baselines comprise PIBT, RHCR/WPPL, LLLG~\citep{arita2026lllg} and
TrafficFlow~\citep{chen2024trafficflow}.

\begin{table}[t]
\centering
{\fontsize{9}{10.8}\selectfont\setlength{\tabcolsep}{3pt}
\begin{tabular*}{\linewidth}{@{\extracolsep{\fill}}L{48pt}C{36pt}L{74pt}L{69pt}L{63pt}L{75pt}@{}}
\toprule
& & \multicolumn{2}{c}{\textbf{Training and reuse}} & \multicolumn{2}{c}{\textbf{Planner involvement}}\\
\cmidrule(lr){3-4}\cmidrule(lr){5-6}
\textbf{Method} & \textbf{Learning} & \textbf{Trained on} & \textbf{Reused on} & \textbf{Training} & \textbf{Execution}\\
\midrule
\rowcolor{ours}
\textbf{Ours} & RL & \shortstack[l]{\rev{Pool of small maps}} & \shortstack[l]{All layouts\\within a class} & None & None \\[5pt]
SILLM & IL & \shortstack[l]{Target map$^{m}$\\(downscaled)} & \shortstack[l]{Larger target;\\new populations} & \shortstack[l]{Planner\\demonstrations} & \shortstack[l]{PIBT; guide paths\\for DG} \\[5pt]
OnlineGGO & ES & \shortstack[l]{Small source\\map$^{m}$} & \shortstack[l]{Larger layouts;\\new populations} & \shortstack[l]{Planner\\rollouts} & \shortstack[l]{PIBT; distance\\updates} \\[5pt]
Follower & RL & \shortstack[l]{40 generated\\mazes} & All layouts & \shortstack[l]{Planned routes\\and rewards} & A* routes \\[5pt]
\bottomrule
\end{tabular*}
\par}
\caption[]{\textbf{Training and reuse \rev{of the learned policies compared in Table~\ref{tab:throughput}}.}
RL, IL and ES denote reinforcement learning, imitation learning and evolution strategies.
$m$ denotes map-specific training; SILLM also uses separate policies for each guidance variant.
The \rev{SILLM comparison} uses IL checkpoints; its authors report that IL outperforms their
MAPPO~\citep{yu2022mappo} implementation~\citep{jiang2025sillm}.
Planner involvement excludes geometric preprocessing.
}
\label{tab:learned_baselines}
\end{table}

All entries in Table~\ref{tab:throughput} are reproduced measurements. Baseline configurations are specified
in Appendix~\ref{app:eval_protocol}, and timing protocols in Appendix~\ref{app:cost}.

\subsection{Throughput and cost across layouts}
\label{sec:results}

We transfer the small-tile policies to larger layouts outside their training pools.
\rev{For each training seed, one terminal checkpoint is reused across all layouts within
its class, without target-layout retraining or checkpoint selection.}
\revbegin{}Goal-distance tables are constructed\revend{} for each deployment map.

\begin{table}[t]
\centering
{\fontsize{9.5}{11.4}\selectfont\setlength{\tabcolsep}{3.5pt}
\begin{tabular*}{\linewidth}{@{\extracolsep{\fill}}L{61pt}*{6}{C{49pt}}@{}}
\toprule
& \multicolumn{2}{c}{\textbf{Structured}} & \multicolumn{4}{c}{\textbf{Unstructured}}\\
\cmidrule(lr){2-3}\cmidrule(lr){4-7}
\textbf{Method} & \textbf{Warehouse} & \textbf{Sortation} & \textbf{Berlin} & \textbf{Paris} & \shortstack{\textbf{Random}\\\textbf{10\%}} & \shortstack{\textbf{Random}\\\textbf{20\%}}\\
\midrule
\rowcolor{ours}
\shortstack[l]{\textbf{Ours}} & \shortstack[c]{42.31\\\textcolor{cost}{\rev{8.4}}} & \shortstack[c]{44.20\\\textcolor{cost}{\rev{7.7}}} & \shortstack[c]{32.99\\\textcolor{cost}{\rev{8.7}}} & \shortstack[c]{33.17\\\textcolor{cost}{\rev{8.7}}} & \shortstack[c]{54.19\\\textcolor{cost}{\rev{5.8}}} & \shortstack[c]{47.52\\\textcolor{cost}{\rev{7.2}}} \\[3pt]
\shortstack[l]{SILLM-SG} & \shortstack[c]{42.25\\\textcolor{cost}{\rev{13.1}}} & \shortstack[c]{43.98\\\textcolor{cost}{\rev{12.9}}} & \shortstack[c]{21.62\\\textcolor{cost}{\rev{13.0}}} & \shortstack[c]{22.53\\\textcolor{cost}{\rev{13.0}}} & \shortstack[c]{52.49\\\textcolor{cost}{\rev{12.9}}} & \shortstack[c]{32.53\\\textcolor{cost}{\rev{13.0}}} \\[3pt]
\shortstack[l]{SILLM-DG} & \shortstack[c]{35.40\\\textcolor{cost}{\rev{277.1}}} & \shortstack[c]{39.39\\\textcolor{cost}{\rev{391.0}}} & \shortstack[c]{30.02\\\textcolor{cost}{\rev{299.8}}} & \shortstack[c]{30.89\\\textcolor{cost}{\rev{311.4}}} & \shortstack[c]{50.50\\\textcolor{cost}{\rev{443.2}}} & \shortstack[c]{44.15\\\textcolor{cost}{\rev{364.3}}} \\[3pt]
\shortstack[l]{SILLM-BD} & \shortstack[c]{27.27\\\textcolor{cost}{\rev{13.0}}} & \shortstack[c]{42.91\\\textcolor{cost}{\rev{12.9}}} & \shortstack[c]{19.16\\\textcolor{cost}{\rev{13.0}}} & \shortstack[c]{18.68\\\textcolor{cost}{\rev{13.1}}} & \shortstack[c]{53.73\\\textcolor{cost}{\rev{12.9}}} & \shortstack[c]{44.09\\\textcolor{cost}{\rev{12.9}}} \\[3pt]
\shortstack[l]{OnlineGGO} & \shortstack[c]{\rev{32.22}\\\textcolor{cost}{\rev{2,419.7}}} & \shortstack[c]{38.13\\\textcolor{cost}{\rev{3,424.7}}} & \shortstack[c]{26.15\\\textcolor{cost}{\rev{3,087.4}}} & \shortstack[c]{27.93\\\textcolor{cost}{\rev{2,809.0}}} & \shortstack[c]{50.15\\\textcolor{cost}{\rev{3,684.8}}} & \shortstack[c]{41.72\\\textcolor{cost}{\rev{3,164.0}}} \\[3pt]
\shortstack[l]{Follower} & \shortstack[c]{8.52\\\textcolor{cost}{\rev{2,629.4}}} & \shortstack[c]{27.14\\\textcolor{cost}{\rev{2,483.5}}} & \shortstack[c]{8.78\\\textcolor{cost}{\rev{3,903.0}}} & \shortstack[c]{10.08\\\textcolor{cost}{\rev{3,372.1}}} & \shortstack[c]{40.34\\\textcolor{cost}{\rev{2,491.4}}} & \shortstack[c]{26.53\\\textcolor{cost}{\rev{2,528.1}}} \\[3pt]
\midrule
\shortstack[l]{WPPL} & \shortstack[c]{42.79\\\textcolor{cost}{\rev{1,808.8}}} & \shortstack[c]{44.25\\\textcolor{cost}{\rev{1,314.6}}} & \shortstack[c]{19.86\\\textcolor{cost}{\rev{1,186.7}}} & \shortstack[c]{22.23\\\textcolor{cost}{\rev{1,502.2}}} & \shortstack[c]{53.45\\\textcolor{cost}{\rev{1,530.2}}} & \shortstack[c]{34.28\\\textcolor{cost}{\rev{1,506.3}}} \\[3pt]
\shortstack[l]{TrafficFlow} & \shortstack[c]{27.45\\\textcolor{cost}{\rev{199.1}}} & \shortstack[c]{36.98\\\textcolor{cost}{\rev{316.9}}} & \shortstack[c]{27.12\\\textcolor{cost}{\rev{229.3}}} & \shortstack[c]{27.10\\\textcolor{cost}{\rev{229.7}}} & \shortstack[c]{46.94\\\textcolor{cost}{\rev{368.7}}} & \shortstack[c]{39.62\\\textcolor{cost}{\rev{286.1}}} \\[3pt]
\shortstack[l]{PIBT} & \shortstack[c]{19.07\\\textcolor{cost}{\rev{7.4}}} & \shortstack[c]{32.48\\\textcolor{cost}{\rev{7.0}}} & \shortstack[c]{14.92\\\textcolor{cost}{7.0}} & \shortstack[c]{15.39\\\textcolor{cost}{7.0}} & \shortstack[c]{46.52\\\textcolor{cost}{\rev{6.8}}} & \shortstack[c]{30.19\\\textcolor{cost}{7.0}} \\[3pt]
\shortstack[l]{LLLG} & \shortstack[c]{17.57\\\textcolor{cost}{\rev{1,470.6}}} & \shortstack[c]{42.33\\\textcolor{cost}{\rev{1,568.9}}} & \shortstack[c]{19.53\\\textcolor{cost}{\rev{2,116.6}}} & \shortstack[c]{20.28\\\textcolor{cost}{\rev{2,105.8}}} & \shortstack[c]{54.80\\\textcolor{cost}{\rev{1,868.5}}} & \shortstack[c]{45.90\\\textcolor{cost}{\rev{1,882.8}}} \\[3pt]
\bottomrule
\end{tabular*}
\par}
\caption[]{\textbf{Throughput and execution cost at 10,000 agents.}
Each cell gives throughput (goals/step) above \textcolor{cost}{step time (ms)}.
Random percentages denote obstacle density. WPPL uses 40,000 LNS iterations.
Ours reuses one policy per class without retraining; throughput averages three training
seeds and eight cases. Ours runs 144 plain allocation iterations on the structured and random layouts
and 288 on the city layouts.
Timing protocols: \papertimingref.
}
\label{tab:throughput}
\end{table}

On warehouse and sortation (Table~\ref{tab:throughput}), the policy trained on the structured \revbegin{}pool\revend{}
matches SILLM-SG's throughput
and stays within 1.3\% of the reproduced WPPL scores, with a
\revbegin{}$1.6$--$1.7\times$\revend{} execution speedup over SILLM-SG.

On unstructured layouts, the policy takes \revbegin{}5.8--8.7\revend{} ms per step. On the 20\% obstacle-density
random map, throughput exceeds SILLM-DG by 7.6\% and
TrafficFlow by 20.0\%, with approximately \rev{$50\times$} and \revbegin{}$39\times$\revend{} faster
execution, respectively. On Berlin and Paris, the same policy also exceeds
SILLM-DG's throughput despite receiving no training crops from either city.

\rev{Across the six benchmark layouts, the transferred policies match or exceed the
best throughput among the evaluated SILLM variants while running faster than
SILLM-SG on every layout.}

Figure~\ref{fig:throughput-cost} also reports throughput and cost at smaller map sizes and populations.
Additional large-layout\rev{, small-map and training-pool} comparisons, per-seed scores and
\num{10000}-step evaluations appear in
Appendix~\ref{app:additional_experiments}.
Step times exclude map preprocessing; Appendix~\ref{app:cost} gives the measurement protocols
and setup costs.

\subsection{Transfer across populations}
\label{sec:density}
\label{sec:cost}

We vary population on three fixed random maps of sizes $32^2$, $64^2$ and $256^2$,
reusing the unstructured policy with unchanged learned parameters (Figure~\ref{fig:density}).
Occupancy denotes agents per free cell; Appendix~\ref{app:density} specifies the evaluation.
The \num{16000}--\num{20000}-agent settings extend beyond the \num{12000}-agent
evaluations reported in the original SILLM and TrafficFlow
studies~\citep{jiang2025sillm,chen2024trafficflow}.

\begin{figure}[t]
\centering
\includegraphics[width=\linewidth]{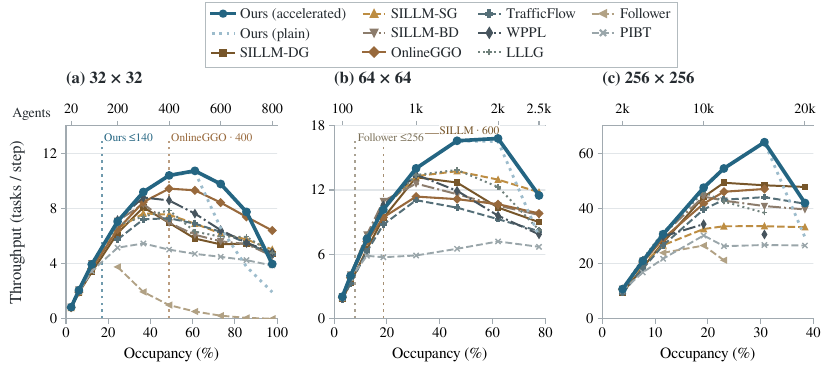}
\caption[Population transfer across three random maps]{\protect\input{figures/density_accelerated_corrected_20260926/caption.tex}}
\label{fig:density}
\end{figure}

Although training uses at most 140 agents, mean throughput exceeds the displayed baselines
at 600 and 700 agents on $32^2$, \num{1000}--\num{2000} on $64^2$, and \num{16000} on $256^2$.
\rev{In extremely dense settings our policy declines abruptly, although a larger
allocation budget recovers part of the loss. We hypothesise the following cause.
Agents that demand the same cell are linked through its destination potential in
Eq.~\eqref{eq:prices}. At such densities these links form groups that extend across
large parts of the map. The fixed iteration budget is then too small to enforce the
cell capacities in Eq.~\eqref{eq:allocation_constraints} throughout a group, so that
agents choose incompatible moves, which the default collision rules cancel. Even with a
converged allocation, agents select their moves independently, and many moves within such
a group remain incompatible.
Appendix~\ref{app:sinkhorn_eval} compares the extent of these groups in training and
evaluation. To address the cancelled moves, it introduces iterative action selection,
which lets these agents choose again and raises throughput further.}

\subsection{Evaluation-time Sinkhorn}
\label{sec:sinkhorn_eval}

\rev{We vary the number and form of Sinkhorn updates at evaluation while keeping the
trained policy parameters fixed.} Temperature scaling~\citep{schmitzer2019stabilized}
starts above the learned temperature $\tau_p$ and decreases to it; over-relaxation
\citep{thibault2021overrelaxed} accelerates the potential updates. Both retain the same
fixed-point equations at the final temperature.

On \rev{Berlin and Paris}, acceleration improves throughput
with half as many iterations, reducing per-step time by about 35\%.
Appendix~\ref{app:sinkhorn_eval} reports convergence, throughput and cost measurements.
Section~\ref{sec:allocation_ablation} complements these evaluations with a
retrained control that removes allocation.

\FloatBarrier
\section{Mechanism and interpretation}
\label{sec:mechanism}
\suppressfloats[t]

We examine how route continuation and destination allocation affect learning and throughput,
using retrained controls matched to each policy's training pool.

\paragraph{Continuation depth.}
The next-move probabilities depend on field scores along the route continuations.
Policy gradients therefore update scores at successive cells along these routes. We examine this
structure by training with $H=1,2,4$, retaining the structured training pool,
training budget and destination allocation. Figure~\ref{fig:mechanisms}a shows
that longer continuations improve validation throughput during training. The terminal
averaged policies retain this ordering on the \num{10000}-agent warehouse.

\begin{figure}[!htbp]
\centering
\includegraphics[width=\linewidth]{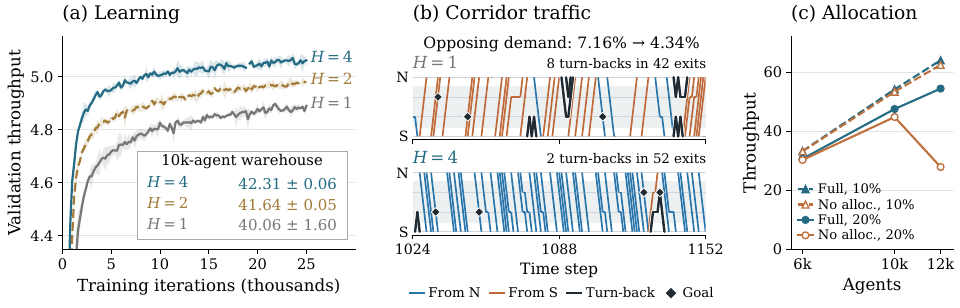}
\caption{\textbf{Continuation and destination allocation.}
(a) Validation throughput with current weights; annotations report terminal
averaged-policy throughput on the \num{10000}-agent warehouse. Both show
three-run means and standard deviations. The deployment annotations average eight
warehouse instances per run.
(b) Actual agent trajectories through one passage; colours indicate entry side
and diamonds mark goal completions. \rev{Thick dark lines mark turn-backs,
agents that leave the passage by the side they entered (12.3\% of exits with
$H=1$ and 7.1\% with $H=4$). The percentages} aggregate all
passages, eight instances and three training runs at 400 agents.
(c) Throughput with and without allocation on 10\% and 20\% obstacle-density random maps,
averaged over three training runs and eight instances per run.}
\label{fig:mechanisms}
\end{figure}

To examine the resulting traffic, we \revbegin{}evaluate the terminal policies\revend{} on a
$33\times57$ warehouse with 400 agents. Across its narrow passages, the mean
probability that neighbouring agents request moves towards each other falls
from 7.16\% with $H=1$ to 4.34\% with $H=4$, while throughput rises from 8.38
to 9.17. Figure~\ref{fig:mechanisms}b \revbegin{}shows one fixed passage. With $H=1$,
agents enter from both ends and turn back when they meet opposing traffic. With $H=4$,
the traffic runs almost entirely in one direction.\revend{}
Preferred travel directions through a passage can change with the traffic state.
These measurements describe \revbegin{}the traffic that each policy generates over time.
We hypothesise that longer continuations improve warehouse throughput by helping
such a one-way convention to form.\revend{} Appendix~\ref{app:continuation_control} gives the protocol,
a comparison on identical traffic states and additional depth comparisons under load.

\paragraph{Destination allocation.}
\label{sec:allocation_ablation}
To test the contribution of allocation, we replace both allocation passes and the
reverse-move penalty with independent normalisation, then retrain with the same
convolutional network, route operator and simulator
(Figure~\ref{fig:mechanisms}c). At \num{12000} agents, removing allocation reduces throughput
from 54.47 to 27.85 on the 20\%-obstacle random map, but by only 2.4\% on the
10\%-obstacle map. Allocation therefore contributes most on the more obstructed layout.
Appendix~\ref{app:allocation_control} reports the full population comparison.

\section{Related work}
\label{sec:related}

\paragraph{Large-scale lifelong MAPF.}
Planning methods balance coordination over future steps against the computation available for
each decision. RHCR solves windowed MAPF problems and reports lifelong experiments with up to
\num{1000} agents~\citep{li2021rhcr}. PIBT uses priority inheritance and backtracking for
single-step coordination, with runtime experiments reaching \num{10000} agents~\citep{okumura2022pibt}.
WPPL refines PIBT-initialized plans using large-neighborhood search and evaluates \num{10000}
agents in lifelong MAPF with rotations~\citep{jiang2024wppl}. Traffic Flow Optimisation supplies
congestion-aware guide paths to PIBT, reporting lifelong experiments with \num{12000}
agents~\citep{chen2024trafficflow}. LLLG combines local guidance with LaCAM in a receding-horizon
planner, evaluated with up to \num{10000} agents~\citep{arita2026lllg}.

PRIMAL and PRIMAL2 combine reinforcement and imitation learning, reporting evaluations with
up to \num{1024} agents in one-shot MAPF and \num{2048} in lifelong MAPF,
respectively~\citep{sartoretti2019primal,damani2021primal2}. PRIMAL2 introduces corridor observations
and movement conventions. Follower learns to follow planned paths and reports lifelong evaluations
with up to 256 agents~\citep{skrynnik2024follower}.
SILLM combines imitation from windowed plan refinement, spatial communication, global guidance
and PIBT-style collision resolution, with population sweeps reaching \num{12000} agents~\citep{jiang2025sillm}.

\paragraph{Spatial guidance and shared representations.}
Spatial guidance and our shared field both associate movement preferences with map locations or
directed moves. Direction maps~\citep{jansen2008directionmaps} and highways~\citep{cohen2015highways}
encode preferred traffic directions. \rev{In pedestrian dynamics, floor-field models draw each walker's
next cell from a static field of distances to the exit and a dynamic field left by other
walkers~\citep{burstedde2001floorfield,kirchner2002bionics}.} GGO \revbegin{}optimises a guidance graph before execution,\revend{} with lifelong
evaluations reaching \num{6000} agents~\citep{zhang2024ggo}.
Online GGO \revbegin{}updates the graph during execution from\revend{} recent traffic and reports \num{12000}-agent
evaluations with guided PIBT~\citep{zang2025onlineggo}.
\rev{In both, the guidance is a set of edge costs for a separate planner through which the agents
act, and the costs or the network producing them are optimised by CMA-ES~\citep{hansen2016cmaes},
which scores each candidate by simulating that planner. Our field is instead a latent \rev{representation} of a
differentiable policy. It is read by the route operator and the allocation and trained by
reinforcement learning from throughput, with no planner in training or execution
(Table~\ref{tab:learned_baselines}).}
SJRL learns environment guidance and communicating subgoal policies, with Causal PIBT resolving
subgoal conflicts; its lifelong experiments with rotations reach 320 agents~\citep{jiang2026sjrl}.
RAILGUN predicts per-cell action distributions with a map-based convolutional policy trained
on planner-generated trajectories, with reported city-tile experiments using up to 256 agents~\citep{tang2025railgun}.
\rev{Its architecture is the closest to ours, with a single convolutional network over the whole map.
It imitates a one-shot MAPF planner and encodes agents by integer index and goals by unnormalised
path cost, whereas our policy learns from its own lifelong experience and uses normalised inputs.
We evaluated its released model on our \num{10000}-agent layouts, where it completes almost no
tasks (Appendix~\ref{app:additional_layouts}), and therefore omit it from the main experiments.}

\revbegin
\paragraph{Probabilistic route choice.}
The reference of Section~\ref{sec:prelim} and the route operator follow models of probabilistic route
choice. Logit route choice in its Markov form describes a route as a sequence of link choices, each made
from the utility of the link and a log-sum value of the remaining trip, so that routes need not be
enumerated~\citep{akamatsu1996cyclic,fosgerau2013rl}. Randomized shortest paths reweight the path
probabilities of a reference random walk exponentially by path cost under a relative-entropy
penalty~\citep{saerens2009rsp,kivimaki2014rsp}, linearly solvable control gives the same form for
controlled Markov chains~\citep{todorov2006linearly}, and maximum-entropy inverse reinforcement learning
fits it to observed routes~\citep{ziebart2008maxent}. In these models the utilities or costs are given with
the network, estimated from observed routes, or tied to the flows by congestion functions. Here they are
computed at every timestep by one network for all agents and apply only to the next $H$ departures of a
fixed reference that already leads to the goal, so no distance or route is recomputed when they change.
\revend

\revbegin
\paragraph{Differentiable allocation.}
Sinkhorn layers provide continuous relaxations for learning latent
permutations~\citep{mena2018gumbelsinkhorn}. Sinkhorn Policy Gradient uses such a
layer in an actor--critic policy over permutations~\citep{emami2018learning}.
Anonymous MAPF has been formulated as an optimal transport problem whose entropy-regularised form
is solved by Sinkhorn iterations~\citep{khan2026mapfot}.
Here, sparse allocation coordinates next-move proposals within a navigation
policy trained from throughput.
\revend

\section{Conclusion}

We presented an LMAPF policy that reuses neural computation across agents through a
shared spatial field. Trained with average-reward reinforcement learning without
planner demonstrations, policies learned on small-map \revbegin{}pools\revend{} transfer to larger,
unseen layouts and populations. Across six layouts with \num{10000} agents, they
attain throughput comparable to or higher than the highest-throughput evaluated
baseline, with execution below 10~ms per step.

\rev{However, there are limitations. We do not address fairness across agents
(Appendix~\ref{app:per_agent}) and leave a fairness-aware extension of our
reinforcement learning formulation to future
work~\citep{jiang2019fairness,zimmer2021fair,asano2023fairdelay}. Our policy also
degrades in extremely dense settings (Section~\ref{sec:density}), where a more
sophisticated allocation strategy may be required
(Appendix~\ref{app:sinkhorn_eval}). Other future directions include assigning tasks
to agents jointly with navigation~\citep{ma2017lifelong} and studying shared spatial
fields in MAPF with
rotations~\citep{jiang2024wppl,jiang2026sjrl,yukhnevich2026epibt}.}

\FloatBarrier
\bibliographystyle{iclr2027_conference}
\subsection*{AI use statement}
We used AI coding assistants to write code to launch experiments and collect data, to search for
related work, and to draft and polish text.

\bibliography{references}
\appendix
\clearpage
\section{Actor specification}
\label{app:actor}

\subsection{Reference movement probabilities}
\label{app:reference}
\label{sec:prior}

\paragraph{Soft-distance construction.}
For a reachable goal $g$, let $\mathcal W_g(x)$ contain finite walks from $x$ whose first visit to $g$ is
their final cell. These walks use valid movement actions, may revisit other cells, and contain no WAIT actions;
$\mathcal W_g(g)$ contains only the empty walk. The ideal soft distance is
\begin{equation}
 D_g(x)=-\frac1\beta\log\sum_{w\in\mathcal W_g(x)}e^{-\beta|w|},\qquad D_g(g)=0.
 \label{eq:softdist}
\end{equation}
We use $\beta=3$. Since the grid degree is at most four and $4e^{-\beta}<1$, this sum converges.
For $x\ne g$, it satisfies
$e^{-\beta D_g(x)}=\sum_{a\in\mathcal M(x)}e^{-\beta(1+D_g(x_a))}$.

\paragraph{Numerical tables.}
Write $\widehat D_g$ for the numerical table used by the actor. The standard builder performs
$2\max(L_y,L_x)+64$ soft-Bellman sweeps in float32 on a map of height $L_y$ and width $L_x$.
The deployment configuration uses a layered fixed-point iteration in
$m_g(x)=\exp\{\beta[d(x,g)-D_g(x)]\}$, where $d$ is shortest-path distance.
It stops when the largest change relative to the updated $m_g(x)$ is below $10^{-3}$
over reachable cells, with a budget of 100 passes, and stores distances in uint16 increments of $1/64$. These finite computations approximate Eq.~\eqref{eq:softdist}.
Tables depend only on the static map and goal,
are constructed before online action selection, and have no learned parameters.

\paragraph{Normalised action rows and goal phase.}
At a live route state $z=(x,\varphi)$, the active goal $g_i(z)$ and
Eq.~\eqref{eq:reference_action} define the reference row. We use the shorthand
\begin{equation}
 p_i^0(a\mid z)=p^0(a\mid x,g_i(z)).
 \label{eq:prior}
\end{equation}
Equation~\eqref{eq:reference_action} normalises rows even for approximate distance tables and adds WAIT separately
from the walks in Eq.~\eqref{eq:softdist}. Before each row is read, arrival at the current goal
promotes the queued goal without \rev{using} an extra departure. Completing both goals makes the
finite continuation absorbing, with a probability-one WAIT row and zero local scores thereafter. Cells with
no valid moves and inactive padding also use WAIT-only rows. Every WAIT \rev{uses} one of the $H=4$
departures. This boundary applies only to the finite continuation; the environment continues assigning tasks.

\subsection{Observation and convolutional network}
\label{app:writer}

The network's raw output is
\begin{equation}
 \Phi_t=\big(\widehat F_{t,0:3},G_t\big),\qquad
 \widehat F_{t,h}(x)\in\mathbb R^5,\quad G_t(x)\in\mathbb R^2.
 \label{eq:field}
\end{equation}
The five-entry rows use $[\mathrm{WAIT},N,S,W,E]$ coordinates, and $G_t$ is the vector branch output.
The shared score $b_h$ and gain inputs $\sigma_h$ in Eq.~\eqref{eq:routegain} are derived from these outputs,
not separate learned heads.

\paragraph{Scene image.}
The convolutional network's 21 input planes consist of occupancy (one plane), the previous realised move of each occupant
(two), the navigable mask (one), descent directions for both goals (eight),
a relative goal-distance feature (one), and rendered reference routes (eight). Previous moves are zero
after a stay. Descent directions and the distance feature are deposited at the agent's current cell. In
particular, the distance feature for agent $i$ is
\[
 \frac{d(x_i,g_i^{\mathrm{next}})}{d(x_i,g_i^{\mathrm{next}})+d(x_i,g_i)+1},
\]
where $d$ is graph distance. Reference-route rendering draws one 40-action route per agent \revbegin{}by
inverse-CDF sampling from the reference probabilities. Draws use a fixed state-based
hash independent of agent ordering.\revend{} Four planes accumulate
directional move counts at departure cells across agents; four more use linearly decreasing weights
to emphasise earlier route steps. All eight are divided by one plus the $9\times9$ local mean of the
total unweighted counts, then passed through $\tanh$. This trace length is separate from the route
operator's four departures.

Replacing the frozen actor's next-goal input with random goals changes mean
throughput by less than 0.3\% on all six main layouts, over three training seeds
and eight cases per seed. On the Random 30\% and Random 40\% layouts
(Appendix~\ref{app:additional_layouts}), the changes are $+1.91\%$ and $-1.38\%$,
respectively.

\paragraph{Architecture and parameters.}
\label{app:actor_params}
The branches have independent encoders and hidden activations. Each base output sums a local
$3\times3$ head and half the summed responses of an affine multiscale head. Two shared kernel bases
$B_0,B_1$ form kernels $B_0-B_1$, $B_0$, and $B_0+B_1$ at dilations 2, 4, and 8; the local head
supplies dilation 1. The action-row bases have shape $5\times48\times3\times3$.
\revbegin{}Each vector basis has shape $2\times8\times8$: two output channels, eight input
channels and eight non-central spatial\revend{} offsets; the local head supplies the centre.

\begin{center}
\begin{tabular}{@{}p{0.22\linewidth}p{0.57\linewidth}r@{}}
\toprule
Component & Architecture & Parameters \\
\midrule
Action-row branch & Two width-48 $3\times3$ ReLU convolutions; local and affine heads & 36,389 \\
Vector branch & Two width-8 $3\times3$ ReLU convolutions; zero-initialised local and affine heads & 2,506 \\
Continuation & Bias-free $15\times48$ pointwise map & 720 \\
Guidance network & & \textbf{39,615} \\
Route operator & Four coefficients and $\tau_f$ & 5 \\
Allocation & $\tau_p$ & 1 \\
\midrule
Total actor & & \textbf{39,621} \\
\bottomrule
\end{tabular}
\end{center}

Writing $\mathcal H_t$ for the action branch's hidden image, the continuation rows are
\begin{equation}
 \widehat F_{t,0}=\widehat F_t,\qquad
 \widehat F_{t,h}=\widehat F_t+W_h\mathcal H_t,\quad h=1,2,3.
 \label{eq:horizonfield}
\end{equation}
The $W_h$ are slices of the pointwise map. The base row and vector output are shared across depths;
all outputs are constructed before route branching and refreshed at the next real timestep.
The reference tables, recurrence and bound have no learned parameters.

\paragraph{Continuation bound.}
Restrict $\widehat F_{t,h}(x)$ to actions valid at $x$, giving $F_{t,h}(x)$, and write
$\Delta_{t,h}=F_{t,h}-F_{t,0}$ for the corresponding depth correction. Let $\bar F$ and $d_F=F-\bar F$
denote a row's valid-entry mean and mean-free part, including WAIT. The fixed transformation is applied to
the sum of the base row and each continuation correction:
\begin{equation}
 \begin{aligned}
 \mathcal B(F)&=\bar F+\min\!\left(1,\frac{\beta_b\tau_f}{\|d_F\|_2}\right)d_F,
 \qquad \beta_b=3,\\
 F^{\mathrm{cont}}_{t,0}&=F_{t,0},\qquad
 F^{\mathrm{cont}}_{t,h}=\mathcal B\!\left(F_{t,0}+\Delta_{t,h}\right),\quad h=1,2,3.
 \end{aligned}
 \label{eq:bound}
\end{equation}
The scale is one at zero norm, with numerical norm stabilisation. The immediate row and all row means
remain unrestricted. Only the corrected continuation row's mean-free norm is bounded in temperature
units by $\beta_b$. Subsequent goal-conditioned adjustment and vector addition are unbounded, so this
bound does not apply to the full score $U_{i,h}$. The transformation adds no learned parameters.

\subsection{Goal-conditioned guidance coefficients}
\label{app:reader}

\paragraph{Goal-conditioned directional and axis preferences.}
At route state $z=(x,\varphi)$, let $p_i^0(\cdot\mid z)$ be the reference row of Eq.~\eqref{eq:prior}.
Use canonical coordinates $[\mathrm{WAIT},N,S,W,E]$, with unavailable move entries set to zero after mapping
from valid actions. Write $\mathcal M(x)$ for the valid movement actions. First remove WAIT and the uniform mass
over valid moves from the reference row:
\[
 r_{\mathrm{WAIT}}=0,\qquad
 r_a=\mathbf1[a\in\mathcal M(x)]
 \left(p_i^0(a\mid z)-\frac{\sum_{m\in\mathcal M(x)}p_i^0(m\mid z)}{|\mathcal M(x)|}\right).
\]
For a cell without valid moves, set $r=0$. The opposite-direction preference $D_i$ and the
vertical--horizontal preference $B_i$ forming $\psi_i$ in Eq.~\eqref{eq:read} are the rows
\begin{align*}
 D_i(z,\cdot)&=\tfrac12(0,r_N-r_S,r_S-r_N,r_W-r_E,r_E-r_W),\\
 B_i(z,\cdot)&=\tfrac14(r_N+r_S-r_W-r_E)(0,1,1,-1,-1),\\
 \psi_i(z,a)&=\big(D_i(z,a),B_i(z,a)\big)^\top.
\end{align*}
These components satisfy $r=D_i+B_i$ and use the same construction for every agent.

\paragraph{Shared scores and field-controlled gains.}
Suppress timestep $t$ and write $F=F^{\mathrm{cont}}_h(x)$ for the departure row. The shared action score and
the two gains used in Eq.~\eqref{eq:read} are
\begin{equation}
 \begin{aligned}
 b_h(x,a)&=F_a+\rev{\langle e_a,G(x)\rangle},\qquad
 \sigma_h(x)=\begin{pmatrix}F_{\mathrm{WAIT}}\\ \sum_{m\in\mathcal M(x)}F_m\end{pmatrix},\\
 \gamma_h(x)&=K\sigma_h(x),\qquad
 K=\begin{pmatrix}\eta_1-\eta_3&\eta_2-\eta_4\\
                   \eta_1+\eta_3&\eta_2+\eta_4\end{pmatrix}.
 \end{aligned}
 \label{eq:routegain}
\end{equation}
\rev{Here $e_a$ is the unit vector in the direction of move $a$ and $e_{\mathrm{WAIT}}=0$. The second term of
$b_h$ is the projection of $G(x)$ on the direction of the move, so a move along $G(x)$ gains what the
opposite move loses.} The learned coefficients $\eta_1,\ldots,\eta_4$ are shared across agents, cells and
depths with unrestricted signs. The gains multiply
$D_i$ and $B_i$, respectively.
The gain inputs contain $F$ only. The term \rev{$\langle e_a,G(x)\rangle$} contributes to $b_h$ after the goal-conditioned adjustment;
including it in the gain inputs would change the route operator at obstacle boundaries.

\paragraph{Obstacle masks.}
Reference centring uses the mean over valid moves. Form $D_i$ and $B_i$ in canonical, zero-padded
coordinates and discard invalid output actions after combining them: either component can be nonzero
in blocked directions. The combined term need not have zero mean over valid moves; its WAIT entry is zero.

\subsection{Computing the guided route distribution}
\label{app:route_law}

\paragraph{Soft dynamic programming.}
The distribution in Eq.~\eqref{eq:measure} is evaluated by a backward recursion.
Write $\mathcal A(z)$ for the available actions at route state $z$, and $z_a$ for its successor after
action $a$, including any goal-phase change.
Define a soft route value $\mathcal V_{i,h}$ in the same score units as $U_{i,h}$:
\begin{equation}
 \begin{aligned}
 \mathcal V_{i,H}(z)&=0,\\
 \mathcal V_{i,h}(z)&=\tau_f\log\sum_{a\in\mathcal A(z)}p_i^0(a\mid z)
  \exp\!\left(\frac{U_{i,h}(z,a)+\mathcal V_{i,h+1}(z_a)}{\tau_f}\right),
 \qquad h=H-1,\ldots,0.
 \end{aligned}
 \label{eq:backup}
\end{equation}
Here $H=4$, and goal phases follow Appendix~\ref{app:reference}. Absorbing and inactive states have
zero local scores and soft value. The first-action marginal is
\begin{equation}
 q_i(a)=p_i^0(a\mid z_0)
 \exp\!\left(\frac{U_{i,0}(z_0,a)+\mathcal V_{i,1}(z_a)
                       -\mathcal V_{i,0}(z_0)}{\tau_f}\right).
 \label{eq:marginal_recursion}
\end{equation}
The implementation uses $\log\max(p_i^0,10^{-30})$ on positive numerical reference probabilities,
with zero-probability and invalid actions masked separately. The finite route law uses the supplied
numerical reference weights, subject to this floor and floating-point evaluation. The root affinity
equals $\tau_f\log q_i(a)$ up to a per-agent constant removed by allocation row centring.

\paragraph{Score derivative.}
Differentiating the finite route marginal with respect to one local score gives
\begin{equation}
 \frac{\partial\log q_i(a_0)}{\partial U_{i,h}(z,b)}
 =\frac1{\tau_f}\left[
  \Pr_{Q_i}(z_h=z,a_h=b\mid a_0)
  -\Pr_{Q_i}(z_h=z,a_h=b)\right].
 \label{eq:score}
\end{equation}
Field gradients chain through the route operator and continuation transform and accumulate across agents, phases
and depths. Equation~\eqref{eq:score} applies before allocation; proposal gradients also pass through
its final allocation pass.

\subsection{Destination allocation and action selection}
\label{app:projection}

For allocation, express each route-operator output in score units as
\begin{equation}
 A_i(a)=\tau_f\log q_i(a).
 \label{eq:rawaffinity}
\end{equation}

\paragraph{Allocation constraints and slack.}
Let $M=|V|$ be the number of cells, $N$ the number of agents, and
$D=M-N$ the number of dummy rows representing unused cell capacity, termed slack rows. Write
$\mathcal C_i=\{x_i^a:a\in\mathcal A(z_i)\}$, with root state $z_i=(x_i,\varphi_i)$, for agent $i$'s candidate
destinations, including its current cell. The augmented allocation in
Eq.~\eqref{eq:allocation} has one row per agent or slack row and one column
per cell, with feasible set
\begin{equation}
 \mathcal U=\left\{T\in\mathbb R_+^{M\times M}:
 T\mathbf1=\mathbf1,\quad T^\top\mathbf1=\mathbf1,\quad
 T_{ix}=0\ \text{if }i\leq N,\ x\notin\mathcal C_i\right\}.
 \label{eq:allocation_constraints}
\end{equation}
Thus every real agent receives one unit of mass and every cell has unit
capacity; slack fills the unused capacity. With fixed affinity matrix $B$, including
a common score \revbegin{}$b_d=-20$\revend{} for slack rows, the ideal assignment is
\begin{equation}
 T^*(B)=\arg\max_{T\in\mathcal U}
 \big\{\langle T,B\rangle+\tau_p\mathcal H(T)\big\},
 \qquad \mathcal H(T)=-\sum_{j,x}T_{jx}\log T_{jx}.
 \label{eq:allocation}
\end{equation}
We apply a fixed number of Sinkhorn iterations to this balanced entropic transport
formulation on the candidate support~\citep{cuturi2013sinkhorn}.
\revbegin{}When a feasible assignment is positive on every entry that $\mathcal U$ leaves free,
the maximiser has the form $T^*_{jx}=\exp[(B_{jx}+f_j+g_x)/\tau_p]$ on these entries, where
$f_j$ and $g_x$ are potentials of the unit-mass constraint of row $j$ and the unit-capacity
constraint of cell $x$. All agents with $x$ among their candidate destinations share $g_x$.\revend{}

Entropy applies to both agent and slack rows. When $D=0$, there are no slack rows and every cell
receives one unit of real-agent mass.

\paragraph{Finite scaling and reverse-edge repricing.}
The allocation procedure subtracts each valid affinity row's arithmetic mean,
$\bar A_i(a)=A_i(a)-|\mathcal A(z_i)|^{-1}\sum_b A_i(b)$.
Row constants leave the finite allocation procedure unchanged in exact arithmetic. Each pass starts from
zero destination potentials. For scores $B_{ix}$, one sweep computes
\begin{align}
 f_i&=-\tau_p\log\sum_{x\in\mathcal C_i}
             e^{(B_{ix}+g_x)/\tau_p},\qquad
 f_d=-\tau_p\log\sum_{x\in V}e^{(b_d+g_x)/\tau_p},\notag\\
 g_x&=-\tau_p\log\left[
       \sum_{i:x\in\mathcal C_i}e^{(B_{ix}+f_i)/\tau_p}
       +D e^{(b_d+f_d)/\tau_p}\right].
 \label{eq:sparse_scaling}
\end{align}
The sums use log space and sparse candidate gathers and scatters. The common slack affinity $b_d$
cancels in $b_d+f_d$ \revbegin{}under the ordinary row update above;\revend{} slack rows are represented by the scalar $f_d$ and multiplicity $D$, without
materialising the augmented matrix.

The first pass uses $B_{i,x_i^a}=\bar A_i(a)$. Its candidate logits are
normalised within each valid row to give $p_i^{(1)}$. The second pass
uses $B_{i,x_i^a}=\bar A_i(a)-c_i(a)$, where
\begin{equation}
 c_i(a)=
 \begin{cases}
  2p_j^{(1)}(b),& a\ne\mathrm{WAIT},\ x_i^a=x_j,
                      \ x_j^b=x_i,\\
  0,&\text{otherwise}.
 \end{cases}
 \label{eq:reverse_edge_price}
\end{equation}
Here $j$ is the current occupant of the candidate destination and $b$ is
its reverse move. WAIT and vacant destinations incur no surcharge. Each
pass uses 48 sweeps during training; \rev{the evaluations of Table~\ref{tab:throughput}} use 144 on structured and random
layouts and 288 on Berlin and Paris (Appendix~\ref{app:eval_protocol}). The objective
in Eq.~\eqref{eq:allocation} describes each pass with its scores held fixed.

\paragraph{Returned probabilities and gradients.}
From the final pass, the allocation procedure returns valid candidate logits
\begin{equation}
 \ell_i(a)=\frac{\bar A_i(a)-c_i(a)+f_i+g_{x_i^a}}{\tau_p},
 \qquad \pi_i(a)=\operatorname{softmax}_{a}\ell_i(a).
 \label{eq:decoder_logits}
\end{equation}
Row normalisation removes $f_i$ and the subtracted affinity mean. Using
Eq.~\eqref{eq:rawaffinity} and setting $\lambda_x=-g_x$ therefore gives
Eq.~\eqref{eq:prices} at any finite sweep count, up to numerical error.

Each sweep ends with column normalisation of the augmented transport,
while finite row-sum errors can remain. The final row normalisation can
therefore violate the cell capacities.

The temperatures are $\tau_f=\exp(\omega_f)$ and
$\tau_p=\exp(\omega_p)$, both initialised at $0.15$ and exempt from weight
decay. The backward pass
differentiates the final allocation pass, including its dependence
on the affinity rows and $\tau_p$. It holds the first pass's
probabilities fixed when forming the reverse-edge surcharge. Training
uses the log probability of the sampled proposals.

\subsection{Construction of the mechanism illustration}
\label{app:illustration}
Figure~\ref{fig:actor} uses terminal averaged weights \revbegin{}from unstructured training
seed 2.\revend{} Selected route prefixes span four departures,
using the current guidance throughout. Ribbon widths indicate the probability
of traversing each edge at its departure depth; a minimum width preserves
rare alternatives. The guidance effect is the reweighted probability minus
the prior probability at the same edge and departure. Numeric route labels
give first-move probabilities. The allocation views compare the operator output
$q_i$ with the complete output $\pi_i$ of Eq.~\eqref{eq:prices}, including
temperature scaling within allocation. B participates in the calculation;
its candidate destinations are disjoint from A's and C's and are omitted
from the connection diagram.

\FloatBarrier
\section{Training details}
\label{app:training}

\paragraph{Shared settings.}
Table~\ref{tab:training} gives the shared optimisation settings. Each seed uses
\num{25000} updates of $128\times64$ joint environment steps, totalling
\num{204800000} steps. Training takes approximately 12--13 hours per seed on one
RTX 5090, including periodic evaluation.
Advantages use generalised advantage estimation~\citep{schulman2016gae}; the actor
uses the clipped PPO surrogate~\citep{schulman2017ppo} with an adaptive KL penalty
and Adam~\revbegin{}\citep{kingma2015adam}\revend{} with decoupled weight decay~\citep{loshchilov2019adamw}.
The scene likelihood ratio is the product of agents' proposal likelihood ratios. The actor and critic have separate parameters; the critic is discarded at deployment.

\begin{table}[h]
\centering
\small
\begin{tabular}{p{0.22\linewidth}p{0.70\linewidth}}
Component & Setting \\
\midrule
learning setup & average reward, $\gamma=1$, per-map rate; entropy temperature $\tau=0.05$ \\
advantage estimation & GAE: reward $\lambda=0.9$, entropy $\lambda=0.5$ \\
actor update & PPO clip $0.296$; 2 passes; 8 minibatches \\
KL penalty & adaptive coefficient, initial value $1$; target $0.015$ \\
critic update & 4 passes, decoupled; \rev{loss $0.751(\mathrm{MSE}_r+\mathrm{MSE}_e)$;} per-map target normalisation; no weight decay \\
optimiser & Adam, constant learning rate $2\times10^{-4}$; actor weight decay $0.03$, learned temperatures exempt \\
\rev{gradient clipping} & \rev{norm limit 10, applied separately to actor and critic updates} \\
batch & $B=128$ environments $\times$ $T=64$ steps per iteration \\
reported checkpoint & terminal \rev{exponentially averaged} weights; decay $0.9998$, initialised from the initial actor and updated once per iteration \\
allocation & \num{48} Sinkhorn iterations per pass during training \\
\bottomrule
\end{tabular}
\caption{Shared training settings for the two selected policies.}
\label{tab:training}
\end{table}

\paragraph{Reward and entropy advantages.}
AR-EAPO~\citep{choe2025areapo} treats reward and entropy as separate learning signals, each centred by its
estimated long-term average.
For map entry $m$, the two signals are $x_t^r=r_t$ and
$x_t^e=-N^{-1}\sum_i\log\pi_{\mathrm{old},i}(a_{i,t}\mid s_t)$, where $N$ counts active agents. Write $V_t^j=V^j_{\chi,m}(O_t^V)$ for critic output
$j\in\{r,e\}$ and $V_{t+1}^{j,+}$ for its value at the successor before any random reset.
The differential TD residual and GAE recursion are
\begin{equation}
 \begin{aligned}
 \delta_t^j&=x_t^j-\widehat\rho_m^j+V_{t+1}^{j,+}-V_t^j,\\
 \mathcal A_t^j&=\delta_t^j+\lambda_j c_t\mathcal A_{t+1}^j,
 \end{aligned}
 \label{eq:training_streams}
\end{equation}
where $c_t=0$ at a reset and $1$ otherwise. At rollout boundaries, the advantage trace
ends with zero and the critic supplies the successor value.
After each rollout, the zero-initialised rates are updated by
$\widehat\rho_m^j\leftarrow\widehat\rho_m^j+\eta_j(1-\lambda_j)\mathbb E_t[\mathcal A_t^j]$,
with $(\eta_r,\eta_e)=(0.23,0.116)$ and $\mathbb E_t$ the empirical average of unnormalised rollout advantages.
The actor divides each advantage by its map's temporal standard deviation plus $10^{-6}$,
combines them as $\widetilde{\mathcal A}_t^r+\tau\widetilde{\mathcal A}_t^e$ with entropy temperature $\tau$,
and subtracts the minibatch mean.

\paragraph{Critic and value-target normalisation.}
The critic augments the actor's 21 channels with four planes at agents' current cells.
These encode normalised time since the last goal completion and three indicators of whether
the current goal is within 1, 2 or 4 moves. Two $3\times3$ convolutions with 128 channels and tanh activations,
global spatial averaging and a shared two-output linear head produce $g_\chi^j(O_t^V)$.
Per-map statistics give $V^j_{\chi,m}=\sigma_m^j g_\chi^j+\mu_m^j$ for target construction.
Targets $\mathcal A_t^j+V^j_{\chi,m}(O_t^V)$ update running first and second moments with weight
$0.2$, then are normalised using the updated moments and clipped to $[-10,10]$.
Each standard deviation is floored at the larger of $0.01$ and $0.1$ times the cross-map RMS
standard deviation. This follows the value-target normalisation of
PopArt~\citep{vanhasselt2016popart}, but the shared head is not rescaled to preserve outputs
when the statistics change.

\paragraph{KL control.}
The penalty uses the old-to-new categorical KL averaged over padded agent rows; inactive rows
are deterministic waits and contribute zero. Its coefficient is multiplied by $1.5$ above
$1.5$ times the target and divided by $1.5$ below the target divided by $1.5$, with bounds
$[10^{-3},10^3]$. The PPO surrogate uses the joint scene likelihood ratio, not per-agent ratios.

\begin{figure}[tbp]
\centering
\includegraphics[width=\linewidth]{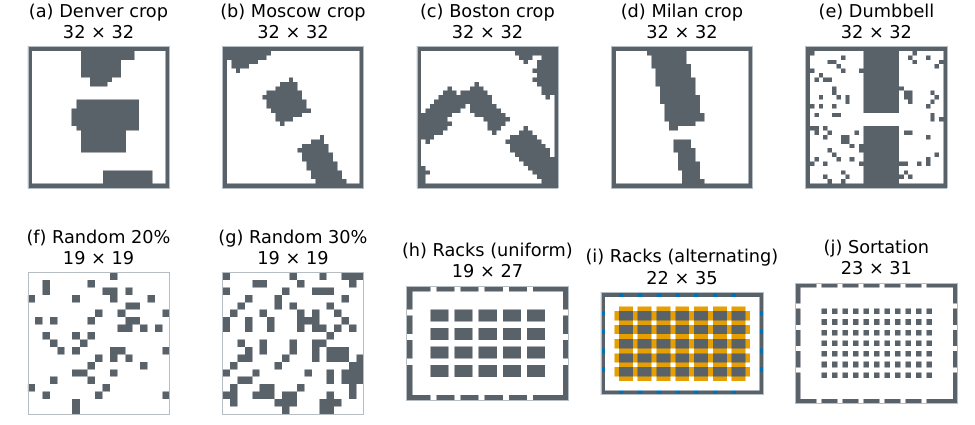}
\caption{Examples from the fixed training pools. Gray cells are obstacles; the dumbbell
layout connects two regions through a narrow passage. In (i), blue and orange mark
workstations and storage endpoints. Dimensions are in cells.}
\label{fig:training-layouts}
\end{figure}

\paragraph{Fixed training pools.}
Figure~\ref{fig:training-layouts} illustrates the training layouts.
Each policy uses three independent training seeds and one fixed pool of \num{128} entries,
shared across its seeds; maps are padded and masked to a common tensor size within the pool.
The structured policy uses 48 rack layouts with uniform goals and \num{128} agents,
48 rack layouts with alternating workstation--storage goals and \num{160} agents, and
32 sortation layouts with uniform goals and \num{128} agents. Rack layouts use a three-cell
outer ring. The unstructured policy uses 48 $32\times32$ street crops and
48 $32\times32$ dumbbell layouts, each with \num{140} agents, plus 16 $19\times19$ random maps
at each of 20\% and 30\% obstacle probability, with \num{64} agents. Dumbbell layouts are rotated
and reflected across the pool. Pool generation uses seeds \num{4300} and \num{10000} for the
structured and unstructured policies, respectively; training and initialisation use seeds 0, 1 and 2.

Street-crop windows are sampled from Denver, Moscow, Boston and Milan in the
Moving AI benchmarks~\citep{sturtevant2012benchmarks}. \revbegin{}After closing each crop's boundary,
valid candidates have 40--85\% free cells with at least 90\% in one connected component,
which is retained. The score is the largest $k\in\{2,3,4,5\}$ for which $k$ cross-shaped
erosions leave at least two components, each containing at least 15\% of the retained
free cells (zero if none). The sampler accepts the first score of at least 3, or the
best valid candidate among 200 proposals.\revend{}

\paragraph{Goal sampling and resets.}
Uniform training goals are drawn from reachable cells at graph distance at least 15 on structured
maps and 10 on unstructured maps. On the alternating-goal rack entries, a workstation source draws
uniformly from storage endpoints; any other source draws uniformly from workstations. Workstations
occupy perimeter gaps, and endpoints are free cells adjacent to interior racks. The distance floor
of 15 applies within the required category; if no candidate satisfies it, the floor relaxes to 1
within that category, preserving alternation. Deployment consumes the released benchmark task streams.
Each environment resets independently at every step with probability $5\times10^{-4}$ on structured
maps or $2.5\times10^{-4}$ on unstructured maps, giving expected intervals of \num{2000} and
\num{4000} steps, respectively. Resets place fresh agents on the same map; states otherwise persist
across updates.

\FloatBarrier
\section{Evaluation protocol}
\label{app:eval_protocol}

\subsection{Layout and small-map comparisons}
\paragraph{Policies and reporting.}
The structured and unstructured policies are trained as specified in
Appendix~\ref{app:training}. Each seed's terminal averaged
parameters are reused across every layout of its class. Throughput counts \revbegin{}the
goals completed over all timesteps;\revend{} we average cases within each seed, then average
seed means equally.

Berlin and Paris are excluded from the training crops.

\paragraph{Baselines.}
The \rev{SILLM comparison of Table~\ref{tab:throughput}} uses released IL checkpoints trained on downscaled target layouts, with
BD, SG and DG guidance as specified in Section~\ref{sec:experiments}. \revbegin{}At the smaller scales
(Figure~\ref{fig:throughput-cost}(a,b) and Table~\ref{tab:middle_comparison}), the SILLM variants use the IL
checkpoints released for Random 20\%.\revend{} Follower
uses its released checkpoint across layouts. The Follower scores reported by
\citet[Table~III]{jiang2025sillm} use target-retrained models and are not paired
with our released-checkpoint timings. The \rev{OnlineGGO on+PIBT comparison of Table~\ref{tab:throughput}}
transfers policies trained on smaller source maps to the corresponding large layouts.
\rev{We changed OnlineGGO's released code in two places. Its traffic observations, which are
zero in the released code, are computed over the previous 20 steps, as its paper describes;
and the 0.5-second limit on its planner is removed in the \num{10000}-agent evaluations.}
TrafficFlow uses GP-R100-Re10-F2
without action-LNS. The \rev{WPPL comparison of Table~\ref{tab:throughput}} uses PIBT-LNS with
\num{40000} iterations, eight threads and a 15-step planning window.
\revbegin{}At \num{10000} agents, LLLG uses planning window 8, horizon 5, two refinements
and a 10-second planning-call limit.\revend{}

\begin{figure}[tbp]
\centering
\includegraphics[width=\linewidth]{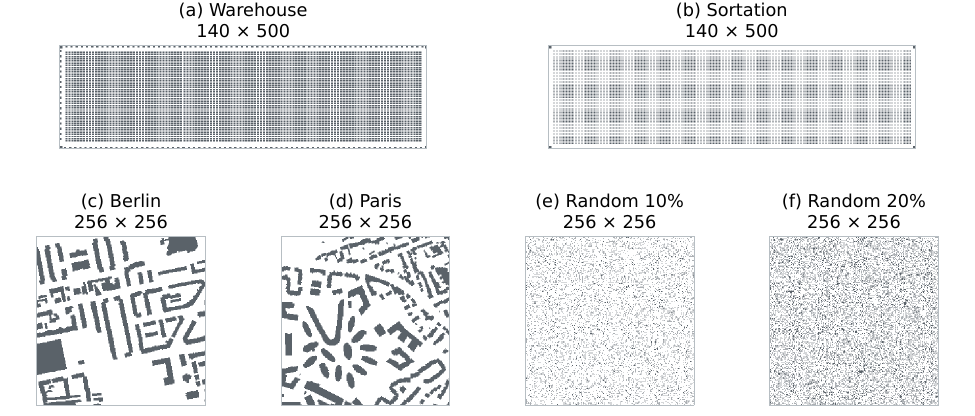}
\caption{The six deployment layouts evaluated with \num{10000} agents. Gray cells
are obstacles; dimensions are in cells.}
\label{fig:deployment-layouts}
\end{figure}

\paragraph{Maps and cases.}
Figure~\ref{fig:deployment-layouts} shows the six large layouts from SILLM~\citep{jiang2025sillm}.
We use its released start/task cases; Table~\ref{tab:protocol} gives the evaluation settings. Random percentages
specify obstacle density; occupancy is agents per free cell.

\begin{table}[htbp]
\centering
\small
\setlength{\tabcolsep}{3pt}
\begin{tabular}{llrrrr}
\toprule
Class & Layout & Size & Free cells & Occupancy & Steps \\
\midrule
Structured & Warehouse & $140\times500$ & \num{38586} & 25.9\% & 3200 \\
Structured & Sortation & $140\times500$ & \num{54320} & 18.4\% & 3200 \\
\midrule
Unstructured & Berlin & $256\times256$ & \num{46880} & 21.3\% & 2500 \\
Unstructured & Paris & $256\times256$ & \num{47096} & 21.2\% & 2500 \\
Unstructured & Random, 10\% & $256\times256$ & \num{58859} & 17.0\% & 2500 \\
Unstructured & Random, 20\% & $256\times256$ & \num{52090} & 19.2\% & 2500 \\
\bottomrule
\end{tabular}
\caption{Deployment maps at \num{10000} agents, with eight released cases (0--7) per
layout. Size is height $\times$ width; occupancy is agents per free cell.}
\label{tab:protocol}
\end{table}

The small random benchmark uses 128 POGEMA~\citep{skrynnik2025pogema} maps, 17--21 cells per axis,
populations 8, 16, 24, 32, 48 and 64, and \revbegin{}256 timesteps.\revend{} Baselines use the
corresponding cases~\citep{li2021rhcr,arita2026lllg,skrynnik2024mats}.
The dense $64^2$/600-agent comparison uses
eight released start/task cases and 256 steps. These populations support the
small and intermediate panels of Figure~\ref{fig:throughput-cost}; the separate
population-transfer study is specified below. Panel (a) of Figure~\ref{fig:throughput-cost}
uses a connected twin of the 128 maps: each map is regenerated with the POGEMA generator
at the same size and obstacle fraction (within 0.01) until its free cells form one
4-connected component, because SILLM's map loader keeps only the largest component
and 43 of the original maps place agents outside it. Every method in panel (a) is
measured on that set with the same instances. PIBT denotes SILLM's released
implementation throughout, so that PIBT and the SILLM variants share one simulator.

\paragraph{Movement rules.}
Training and evaluation use POGEMA's soft collision rules~\citep{skrynnik2025pogema}.
Attempted swaps are cancelled. At a contested cell, a stationary occupant takes precedence
over incoming agents; otherwise, the lowest-index mover takes precedence. Rejected moves
become WAIT, and the resulting cancellations propagate until no conflicts remain.
\revbegin{}This cancellation process does not reconsider alternative moves.\revend{}

\paragraph{Execution.}
The \rev{policies of Table~\ref{tab:throughput}} use argmax proposals and layered 16-bit reference tables. Table~\ref{tab:throughput} uses
144 plain allocation iterations \revbegin{}per pass\revend{} on structured and random layouts and
288 on Berlin and Paris.
\revbegin{}Throughput evaluations may batch independent instances; execution-cost measurements
use one policy\revend{} on one instance. Every task stream is long enough to supply fresh
goals \revbegin{}until the last timestep.\revend{}

Map preprocessing and execution-cost measurements are reported in Appendix~\ref{app:cost}.

\subsection{Population-transfer study}
\label{app:density}

\paragraph{Policy reuse.}
Figure~\ref{fig:density} uses the same unstructured policy as \rev{Table~\ref{tab:throughput}}, with its training pool specified in Appendix~\ref{app:training}.
Each seed reuses its terminal averaged parameters across every map and population,
with argmax proposals and 288 accelerated allocation iterations, increased to
\num{4608} at \num{20000} agents on the $256^2$ map. A plain
288-iteration reference uses the same checkpoints and cases. The accelerated
updates are specified in Appendix~\ref{app:sinkhorn_eval}. Geometric route tables
are constructed for each deployment map.

\paragraph{Evaluation cases.}
The $32^2$, $64^2$ and $256^2$ maps contain 819, \num{3221} and \num{52090} free
cells, respectively. \revbegin{}Each case runs for \num{1000} timesteps\revend{} on the two smaller maps and
\num{2500} on the largest, \revbegin{}and throughput counts all of them.\revend{}
Our policy uses eight start/task cases at every population. The $32^2$ cases use uniform goals
on OnlineGGO's random map. For the two larger maps, additional populations use
SILLM's released agent-placement generator and task streams.
Our points average three training seeds and eight cases per seed.

\paragraph{Baseline training and execution.}
OnlineGGO is trained on the $32^2$ evaluation map with 400 agents using the
released recipe; its guidance is updated every 20 steps, with the initial
20-step unit-guidance warm-up \revbegin{}counted in the throughput.\revend{}
The SILLM policies are trained on the $64^2$ evaluation map with 600 agents.
Follower uses the released policy trained on 40 separate $65^2$ mazes with
128 or 256 agents. The vertical guides therefore mark training agent counts,
not the occupancy of the source maps. WPPL uses \num{2500}, \num{10000} and
\num{40000} LNS iterations on the $32^2$, $64^2$ and $256^2$ maps, respectively
(Appendix~\ref{app:cost}); TrafficFlow uses GP-R100-Re10-F2 without additional action-LNS.
The PIBT curve uses SILLM's released implementation. Unmeasured populations
are left as gaps in the curves.

\FloatBarrier
\section{Execution-cost protocol and measurements}
\label{app:cost}

\paragraph{Execution cost.}
\revbegin{}Execution cost is wall-clock time per simulated step for one method on one
instance, including inference or planning and simulator advancement. Our policy
timings cover all timesteps, with the GPU synchronised, and exclude setup,
compilation and warm-up. Dedicated measurements\revend{} use an AMD Ryzen 7 7800X3D and RTX 5090.
\revbegin{}All step times in the paper are measured on one such machine, one job
at a time. Table~\ref{tab:throughput} uses the first case of each layout (eight cases for PIBT).
Table~\ref{tab:timingprotocol} summarises the timed\revend{} intervals; throughput always
\revbegin{}counts all timesteps.\revend{}

\begin{table}[htbp]
\centering
\small
\revbegin
\begin{tabular}{@{}p{0.24\linewidth}p{0.71\linewidth}@{}}
Method & Timed interval \\
\midrule
Ours, SILLM-SG/BD/DG & All timesteps. \\
PIBT & Complete simulation loop, excluding initialisation. \\
TrafficFlow & Complete simulation loop. \\
WPPL & All timesteps. \\
LLLG & All timesteps, excluding setup. \\
OnlineGGO & All timesteps. \\
Follower & All timesteps. \\
\bottomrule
\end{tabular}
\caption{Timed intervals for the six \num{10000}-agent layouts in
Table~\ref{tab:throughput}. All intervals include simulator execution.}
\label{tab:timingprotocol}
\end{table}

\paragraph{Hardware and baseline settings.}
\rev{The timings of Table~\ref{tab:throughput}} use 16 OpenMP threads for SILLM-SG/DG, eight for SILLM-BD,
\rev{one for PIBT and TrafficFlow} and eight LNS threads for WPPL.
\revbegin{}Planner configurations are given in Appendix~\ref{app:eval_protocol}.\revend{}

\paragraph{WPPL refinement budget.}
Table~\ref{tab:wppl_budget} varies the LNS iteration budget on Random 10\% with
\num{10000} agents. Increasing the budget from \num{2500} to \num{40000}
improves throughput by 3.0\% while increasing execution cost by \rev{$6.4\times$}.
On the \num{10000}-agent layouts, \num{2500} iterations give 41.97 on the warehouse,
43.75 on sortation (seven of eight cases) and 28.34 on Random 20\%, against 42.79, 44.25
and 34.28 at \num{40000} iterations (Table~\ref{tab:throughput}), at \revbegin{}232 and 268\revend{}~ms per step on the
warehouse and Random 20\% (complete runs on the dedicated hardware).
The population study (Figure~\ref{fig:density}) scales the budget with map
size: \num{2500} iterations on the $32^2$ map, \num{10000} on $64^2$ and
\num{40000} on $256^2$, the setting SILLM's harness applies to the
\num{10000}-agent layouts. At \num{40000} iterations on Random 20\%,
\revbegin{}case~0 costs\revend{} \revbegin{}\num{1012}, \num{1506} and \num{1621}\revend{}~ms per step at
\num{6000}, \num{10000} and \num{12000} agents.
At the smaller budgets, \revbegin{}case~0 costs 46--76~ms\revend{} per step on the $32^2$
map (200--800 agents, \num{2500} iterations) and \rev{103--260~ms} per step on the
$64^2$ map (100--\num{2500} agents, \num{10000} iterations).

\begin{table}[htbp]
\centering
\small
\begin{tabular}{rrr}
\toprule
LNS iterations & Throughput (tasks/step) & Execution cost (ms/step) \\
\midrule
\num{2500} & $51.867 \pm 0.089$ & \rev{239.9} \\
\num{10000} & $52.725 \pm 0.110$ & \rev{448.9} \\
\num{40000} & $53.448 \pm 0.107$ & \rev{\num{1530.2}} \\
\bottomrule
\end{tabular}
\caption{WPPL refinement-budget comparison. Throughput is mean $\pm$ standard
deviation over eight \num{2500}-step cases. Costs use case 0 over all timesteps
on the dedicated hardware above.}
\label{tab:wppl_budget}
\end{table}

\paragraph{Map preparation.}
Reference tables are constructed once per map and \revbegin{}shared across agents and
successive tasks (Table~\ref{tab:mapsetup}). Their storage is already included
in the deployment memory measurements below. Their size grows with the square of the number
of free cells, which limits the map size that fits on one GPU.\revend{}

\begin{table}[htbp]
\centering
\small
\begin{tabular}{lrr}
\toprule
Map & Construction (s) & \rev{Shared tables} (GiB) \\
\midrule
Warehouse & \rev{17.5} & 5.53 \\
Sortation & \rev{26.3} & 10.86 \\
Berlin & \rev{20.2} & 8.13 \\
Paris & \rev{20.5} & 8.21 \\
Random, $10\%$ & \rev{29.1} & 12.70 \\
Random, $20\%$ & \rev{22.7} & 10.00 \\
\bottomrule
\end{tabular}
\caption{Reference-table construction times and sizes on RTX 5090. Times cover the host
construction call; compilation and asynchronous completion are included in warm-up.}
\label{tab:mapsetup}
\end{table}

\paragraph{Deployment memory.}
Table~\ref{tab:deploymemory} \revbegin{}separates preparation from execution memory at
\num{10000} agents, distinguishing setup peaks from memory used during execution.
SILLM-SG/BD and our policy store precomputed map tables on the GPU and have
comparable GPU requirements. SILLM-DG computes guidance on the host and transfers
local values to the GPU for inference. On the host, the evaluated PIBT and WPPL
implementations cache distance tables; WPPL and LLLG also retain candidate paths
and search states for replanning. TrafficFlow stores guiding paths and their
distance estimates, while Online GGO maintains distance estimates under its
learned guidance weights.\revend{}

\begin{table}[htbp]
\centering
\small
\setlength{\tabcolsep}{3.5pt}
\revbegin
\begin{tabular*}{\linewidth}{@{\extracolsep{\fill}}lrrrr@{\hspace{14pt}}rrrr@{}}
\toprule
& \multicolumn{4}{c}{Random 20\%} & \multicolumn{4}{c}{Warehouse} \\
\cmidrule(lr){2-5}\cmidrule(lr){6-9}
& \multicolumn{2}{c}{Host} & \multicolumn{2}{c}{GPU} & \multicolumn{2}{c}{Host} & \multicolumn{2}{c}{GPU} \\
\cmidrule(lr){2-3}\cmidrule(lr){4-5}\cmidrule(lr){6-7}\cmidrule(lr){8-9}
Method & Prep. & Exec. & Prep. & Exec. & Prep. & Exec. & Prep. & Exec. \\
\midrule
Ours & 12.1 & 7.5 & 10.3 & 10.3 & 7.4 & 5.1 & 5.8 & 5.8 \\
SILLM-SG & 20.0 & 1.7 & 10.1 & 10.7 & 11.6 & 1.8 & 5.6 & 6.2 \\
SILLM-DG & 1.0 & 23.3 & $<0.1$ & 0.6 & 0.9 & 22.2 & $<0.1$ & 0.6 \\
SILLM-BD & 20.2 & 1.8 & 10.1 & 10.7 & 9.8 & 1.8 & 5.6 & 6.2 \\
PIBT & 10.4 & 10.6 & \textemdash & \textemdash & 5.9 & 6.0 & \textemdash & \textemdash \\
WPPL & 10.8 & 11.5 & \textemdash & \textemdash & 6.2 & 6.9 & \textemdash & \textemdash \\
TrafficFlow & 0.1 & 23.3 & \textemdash & \textemdash & 0.1 & 22.5 & \textemdash & \textemdash \\
LLLG & 2.2 & 8.4 & \textemdash & \textemdash & 1.6 & 6.5 & \textemdash & \textemdash \\
Online GGO & 4.5 & 41.8 & \textemdash & \textemdash & 4.7 & 47.1 & \textemdash & \textemdash \\
Follower & 30.4 & 31.2 & 0.1 & 0.2 & 32.5 & 33.4 & 0.1 & 0.2 \\
\bottomrule
\end{tabular*}
\caption{Peak host memory and live GPU allocation (GiB) during preparation and
execution. Peaks need not coincide; dashes denote CPU-only methods.}
\label{tab:deploymemory}
\end{table}

\FloatBarrier
\section{Additional evaluation results}
\label{app:additional_experiments}

\subsection{Deployment results by training seed}
Table~\ref{tab:deployment_seeds} gives the seed-level plain-allocation results behind
Table~\ref{tab:throughput}, under the protocol in Appendix~\ref{app:eval_protocol}.

\begin{table}[htbp]
\centering
\small
\begin{tabular}{lrrr}
\toprule
Layout & Seed 0 & Seed 1 & Seed 2 \\
\midrule
Warehouse & 42.272 & 42.387 & 42.279 \\
Sortation & 44.191 & 44.228 & 44.180 \\
\midrule
Berlin & 33.518 & 32.033 & 33.412 \\
Paris & 33.618 & 31.097 & 34.809 \\
Random 10\% & 54.161 & 54.217 & 54.180 \\
Random 20\% & 47.483 & 47.564 & 47.514 \\
\bottomrule
\end{tabular}

\caption{Terminal-checkpoint throughput by training seed; each entry averages eight cases.}
\label{tab:deployment_seeds}
\end{table}

\revbegin
\subsection{Per-agent task completion and waiting}
\label{app:per_agent}
Table~\ref{tab:per_agent} describes individual agents in a repeat of the
evaluations of Table~\ref{tab:throughput} with the same checkpoints and cases,
in which mean throughput agrees within 0.2 tasks per step. Three training
seeds, eight cases and \num{10000} agents give \num{240000} agent records per
layout. For each agent
we record the number of completed tasks and the longest run of consecutive
steps without movement.

\begin{table}[htbp]
\centering
\small
\revbegin
\begin{tabular}{@{}lrrrrrrrr@{}}
\toprule
& \multicolumn{4}{c}{Completed tasks per agent} & & \multicolumn{3}{c}{Longest stop (steps)} \\
\cmidrule(lr){2-5}\cmidrule(l){7-9}
Layout & Mean $\pm$ SD & Min & IQR & Max & No task & Median & IQR & Max \\
\midrule
Warehouse & $13.5\pm2.3$ & 6 & 12--15 & 29 & 0 & 3 & 3--4 & 207 \\
Sortation & $14.1\pm2.4$ & 6 & 12--16 & 28 & 0 & 2 & 2--2 & 252 \\
Berlin & $8.2\pm3.0$ & 0 & 6--10 & 20 & 0.63\% & 10 & 6--18 & \num{2374} \\
Paris & $8.2\pm3.1$ & 0 & 6--10 & 21 & 0.34\% & 9 & 5--20 & \num{2353} \\
Random 10\% & $13.5\pm2.0$ & 6 & 12--15 & 26 & 0 & 3 & 2--5 & 374 \\
Random 20\% & $11.9\pm1.9$ & 0 & 11--13 & 23 & 0.001\% & 6 & 4--9 & \num{2234} \\
\bottomrule
\end{tabular}

\vspace{0.8em}
\begin{tabular}{@{}lrrr@{}}
\toprule
Layout & CV & CV of upper bound & Completed / upper bound \\
\midrule
Warehouse & 0.17 & 0.17 & $0.94\pm0.05$ \\
Sortation & 0.17 & 0.17 & $0.97\pm0.04$ \\
Berlin & 0.37 & 0.15 & $0.62\pm0.21$ \\
Paris & 0.38 & 0.17 & $0.65\pm0.21$ \\
Random 10\% & 0.15 & 0.15 & $0.95\pm0.04$ \\
Random 20\% & 0.16 & 0.15 & $0.85\pm0.06$ \\
\bottomrule
\end{tabular}

\caption{Per-agent task completion and waiting at \num{10000} agents, pooled
over three training seeds and eight cases. SD is the standard deviation and IQR
the interquartile range; \emph{No task} is the share of agents that complete no
task within the evaluation horizon. The lower part compares the completed tasks
with their upper bound, the number of tasks that an agent would complete on
shortest paths with no other agent on the map. CV is the coefficient of
variation, and \emph{Completed / upper bound} is the mean $\pm$ SD over agents
of the ratio of the two.}
\label{tab:per_agent}
\end{table}

On warehouse, sortation and Random 10\%, every agent completes at least six
tasks and the standard deviation is 15--17\% of the mean. Random 20\% is
similar, except that 0.04\% of agents complete fewer than six tasks and three
complete none. On Berlin and Paris the distribution is wider, and 0.6\% and
0.3\% of agents complete no task within \num{2500} steps. Most of these agents
keep moving without reaching their goals; their median longest stop is 32 steps
on Berlin and 15 on Paris. On Berlin, Paris and Random 20\%, the longest stops
exceed \num{2200} of the \num{2500} steps.

\paragraph{Comparison with the assigned tasks.}
The goal sequence of each agent, which the benchmark fixes before the
evaluation, also contributes to this spread, since some sequences need longer
paths than others. We measure this contribution following two existing metrics.
\citet{li2021mapflns} define the delay of a path in one-shot MAPF as the number
of steps it takes beyond the shortest path between its endpoints, and
\citet{asano2023fairdelay} measure fairness in multi-robot navigation by the
variance of a similar delay, taken against the trajectory that an agent could
follow in the absence of other agents. In the same way, we compute for every
agent an upper bound on its completed tasks, the number of tasks it would
complete if it followed a shortest path to each of its goals with no other
agent on the map. The lower part of Table~\ref{tab:per_agent} compares the
coefficient of variation (CV) of the completed tasks, the standard deviation
divided by the mean, which \citet{jiang2019fairness} use to measure fairness,
with the CV of the upper bound. It also gives the ratio of the completed tasks
to the upper bound.

\begin{figure}[htbp]
\centering
\includegraphics[width=\linewidth]{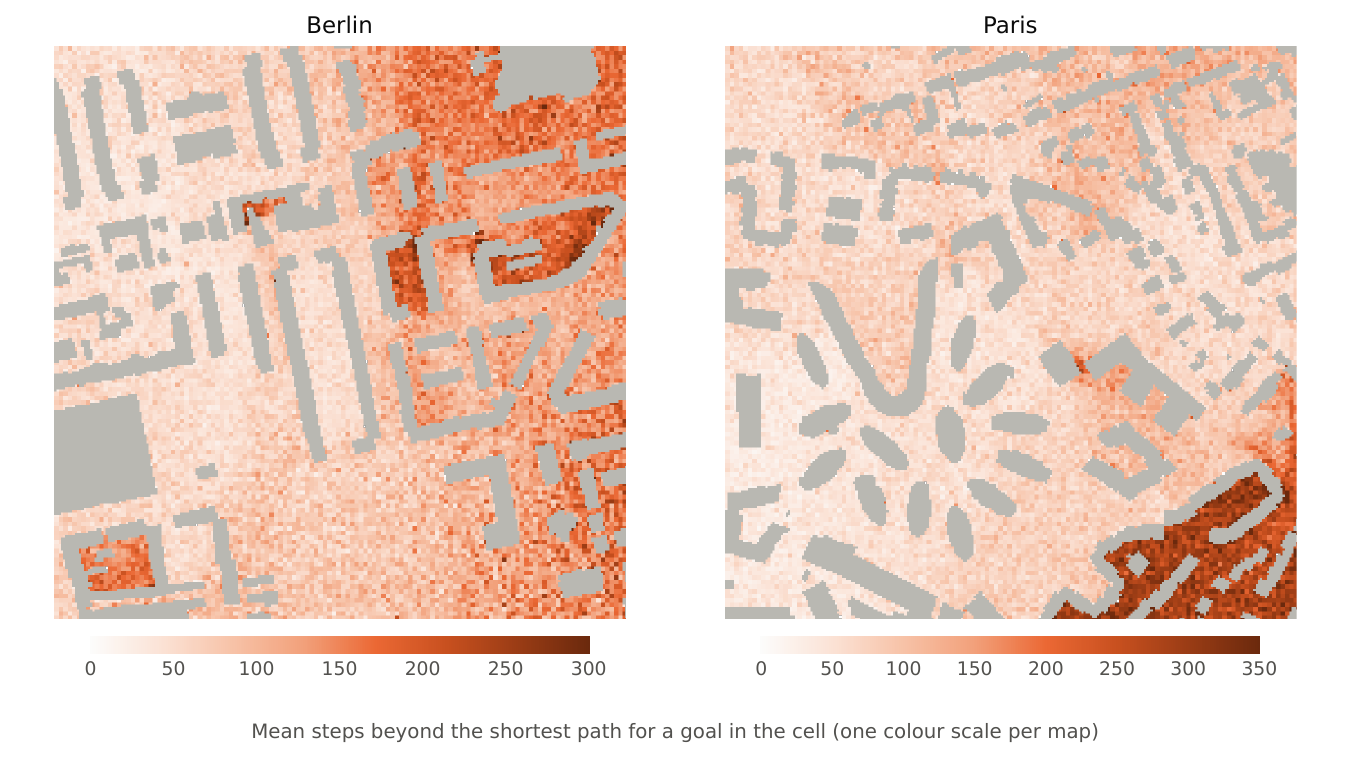}
\revbegin
\caption{Delay by goal location on Berlin and Paris with \num{10000} agents
over \num{2500} steps (Appendix~\ref{app:per_agent}). The delay of a task is
the number of steps it takes beyond the shortest path to its goal. Each cell
shows the mean delay of the tasks whose goal lies in it, taken over blocks of
$2\times2$ cells, three training seeds and eight cases. A task that is
unfinished at the end of the evaluation is counted with its delay up to then.
Each map has its own colour scale. Grey cells are obstacles.}
\label{fig:delay_by_goal}
\end{figure}

On warehouse, sortation and the two random layouts, the CV of the completed
tasks is close to that of the upper bound, and the ratio of the completed tasks
to the upper bound is nearly the same for all agents. On these layouts the
spread comes mainly from the goal sequences. On Berlin and Paris the CV is more
than twice that of the upper bound, and the ratio varies widely between agents.
Figure~\ref{fig:delay_by_goal} shows the delay of the tasks by the location of
their goals, measured in a further repeat of the evaluations that records the
step at which every task is completed. The delay is concentrated in some
regions of each map, which include the regions of congestion described in
Appendix~\ref{app:long_horizon}. We therefore hypothesise that the wider spread
on the two city layouts comes from congestion at particular places, which the
policy does not resolve.

The training objective is aggregate
throughput and contains no fairness term. Fairness-aware objectives and
yielding mechanisms studied in multi-agent reinforcement
learning~\citep{jiang2019fairness,zimmer2021fair,asano2023fairdelay} are a
possible extension.
\revend

\subsection{Small and intermediate operating scales}
Table~\ref{tab:small_all} reports six populations on the \revbegin{}original\revend{} 128-map random
benchmark, using the unstructured policy with 144 plain allocation iterations and
\revbegin{}256 timesteps,\revend{} with RHCR as the reference across populations.

\begin{table}[htbp]
\centering
\small
\begin{tabular}{rrrrr}
\toprule
Agents & Seed 0 & Seed 1 & Seed 2 & RHCR \\
\midrule
8 & 0.607 & 0.610 & 0.607 & 0.625 \\
16 & 1.208 & 1.211 & 1.207 & 1.251 \\
24 & 1.725 & 1.728 & 1.722 & 1.798 \\
32 & 2.175 & 2.174 & 2.164 & 2.286 \\
48 & 3.041 & 3.042 & 3.036 & 3.203 \\
64 & 3.795 & 3.795 & 3.785 & 3.819 \\
\bottomrule
\end{tabular}

\caption{Throughput on 128 small random maps at every tested population.
Policy columns distinguish training seeds; each covers the same benchmark cases as RHCR.}
\label{tab:small_all}
\end{table}

Table~\ref{tab:middle_comparison} compares \revbegin{}methods on\revend{} eight dense $64^2$/600-agent
cases over 256 steps. \revbegin{}At this intermediate scale, our throughput is within
3.3\% of WPPL and\revend{} \revbegin{}LLLG, 6.3\% below SILLM-BD, 0.5\% above SILLM-SG\revend{} and 11.5\% above SILLM-DG.
RHCR uses PBS~\citep{ma2019pbs} with SIPP~\citep{phillips2011sipp}, simulation/planning windows 5/10 and a 10-second planning-call
limit, reached by 46--49 calls per case.

\begin{table}[htbp]
\centering
\small
\begin{tabular}{lr}
\toprule
Method & Throughput \\
\midrule
Ours (three seeds) & 9.778 \\
WPPL & 10.108 \\
LLLG & 10.007 \\
\rev{SILLM-BD} & \rev{10.438} \\
\rev{SILLM-SG} & \rev{9.733} \\
SILLM-DG & 8.766 \\
PIBT & 7.821 \\
RHCR (PBS/SIPP) & 0.892 \\
\bottomrule
\end{tabular}

\caption{Reproduced dense $64^2$/600-agent comparison, eight cases and 256 steps.
Ours uses plain allocation and averages three training seeds.}
\label{tab:middle_comparison}
\end{table}

\FloatBarrier
\subsection[Evaluation-time allocation]{Evaluation-time \rev{allocation}}
\label{app:sinkhorn_eval}
\label{app:population_results}

\revbegin{}Destination allocation supplies the proposal probabilities $\pi_i$ used for
action selection (Section~\ref{sec:projection}). Accelerated updates change how
these probabilities are computed; iterative action selection reuses them to
reconsider moves cancelled by conflicts. Both operate at evaluation with the
same learned field, route operator and trained parameters.\revend{}

\paragraph{Accelerated updates.}
\leavevmode\revbegin{}Destination allocation adjusts the route probabilities $q_i$ for competing
demand for cells (Section~\ref{sec:projection}). In each of its two passes,
plain Sinkhorn iterations alternate row normalisation, which assigns unit mass
to each agent, and destination normalisation, which balances demand against
unit cell capacity (Appendix~\ref{app:projection}). We accelerate these updates
at evaluation\revend{} with temperature scaling~\citep{schmitzer2019stabilized} and
over-relaxation~\citep{thibault2021overrelaxed}. \revbegin{}Temperature scaling begins with
a higher entropy weight and gradually returns to the learned temperature
$\tau_p$; over-relaxation extrapolates each ordinary potential update.\revend{}
For an allocation pass with \revbegin{}$K\geq2$\revend{} iterations, iteration
$k\in\{0,\ldots,K-1\}$ uses
\begin{equation}
 \revbegin{}\tau_{p,k}=\tau_p\,\kappa^{\max\{1-k/\lfloor K/2\rfloor,\,0\}},\revend{}
 \qquad
 \revbegin{}f_i\leftarrow(1-\omega)f_i+\omega\widetilde f_i.\revend{}
 \label{eq:accelerated_scaling}
\end{equation}
Here \revbegin{}$\kappa\geq1$ is the initial-temperature multiplier and $\omega$ is the
relaxation factor. We fix $\kappa=8$ and $\omega=1.6$ in these evaluations;
the temperature reaches $\tau_p$ halfway through each pass and remains there.
The value $\widetilde f_i$ is the ordinary row\revend{} update in
Eq.~\eqref{eq:sparse_scaling}, evaluated at $\tau_{p,k}$; \revbegin{}the same relaxation
applies to the slack potential $f_d$ and destination potentials $g_x$.\revend{}
Row and slack potentials are updated before computing the destination update.
Both allocation passes start from zero potentials and use this schedule;
the reverse-edge repricing rule is unchanged. \revbegin{}Setting $\kappa=\omega=1$
recovers plain scaling.\revend{} At the final temperature $\tau_p$, the relaxed updates
have the same fixed-point equations as plain scaling. \revbegin{}At a finite count, however, their output
is no longer exactly invariant to row constants or to the slack score $b_d$
(Appendix~\ref{app:projection}); this dependence vanishes as the iterations converge.\revend{} Learned parameters
remain fixed, and all checkpoints were trained with 48 plain Sinkhorn
iterations per pass.

\paragraph{Iterative action selection.}
\revbegin{}\label{app:allocation_repair}
At deployment, each agent proposes its most probable action under $\pi_i$
(Section~\ref{sec:projection}). POGEMA's movement rules
(Appendix~\ref{app:eval_protocol}) cancel conflicting moves to WAIT.
Iterative action selection lets agents whose moves were cancelled reconsider
alternatives using the already computed allocation probabilities.\revend{}
\rev{It differs from collision shielding for learned policies, in which
PIBT~\citep{okumura2022pibt} resolves conflicts among each agent's ranked
actions~\citep{veerapaneni2024cspibt} and which SILLM applies to its network
outputs~\citep{jiang2025sillm}. Unlike PIBT, it keeps every move accepted by
the movement rules and uses neither priority inheritance nor backtracking;
only agents whose moves were cancelled choose again, among moves to unassigned cells.}

\revbegin{}First apply the movement rules provisionally and reserve every assigned next
cell, including each waiting agent's current cell. In each round, agents whose
original moves were cancelled and who still wait choose, in parallel, their
highest-scoring action among WAIT and valid moves to unassigned cells.
Comparisons use the fixed logits in Eq.~\eqref{eq:decoder_logits} and the
original action order for ties. For each proposed non-WAIT destination, the
lowest-index proposer is accepted; its destination stays reserved and its
current cell becomes available in the next round. This preserves feasibility
while permanently adding moves. Repeat until a round accepts no additional
move, then advance the simulator once with the resulting joint proposal.
The learned field, route probabilities and Sinkhorn allocation are each
computed only once per timestep.
In the long-horizon comparison of Figure~\ref{fig:long-horizon-six-maps},
iterative selection changes mean throughput by less than 0.5\% on warehouse,
sortation and both random layouts, while improving it on Berlin and Paris.\revend{}

\paragraph{Deployment results.}
\revbegin{}We first evaluate accelerated updates with independent argmax action selection.
The population comparisons below and the long-horizon evaluation in
Appendix~\ref{app:long_horizon} also evaluate iterative action selection.\revend{}
Table~\ref{tab:sinkhorn_deployment} compares plain allocation with 144 accelerated
iterations on the six \rev{layouts of Table~\ref{tab:throughput}}. Throughput averages eight cases within
each of three training seeds. Execution cost uses seed~0 \revbegin{}and case~0,
timed as\revend{} in Appendix~\ref{app:cost}.

\begin{table}[htbp]
\centering
\small
\begin{tabular}{lrrrrr}
\toprule
& \multicolumn{3}{c}{Plain} & \multicolumn{2}{c}{Accelerated, 144 iterations} \\
\cmidrule(lr){2-4}\cmidrule(lr){5-6}
Layout & Iterations & Throughput & ms/step & Throughput & ms/step \\
\midrule
Warehouse & 144 & 42.31 & \rev{8.37} & 42.30 & \rev{8.47} \\
Sortation & 144 & 44.20 & \rev{7.67} & 44.19 & \rev{7.78} \\
Berlin & 288 & 32.99 & \rev{8.69} & 33.47 & \rev{5.64} \\
Paris & 288 & 33.17 & \rev{8.72} & 33.79 & \rev{5.67} \\
Random 10\% & 144 & 54.19 & \rev{5.85} & 54.19 & \rev{6.01} \\
Random 20\% & 144 & 47.52 & \rev{7.25} & 47.53 & \rev{7.62} \\
\bottomrule
\end{tabular}

\caption{Evaluation-time allocation at \num{10000} agents. Throughput is in
tasks per step and execution time in ms per step. Plain values are those of
Table~\ref{tab:throughput}; accelerated values use 144 iterations on every layout.}
\label{tab:sinkhorn_deployment}
\end{table}

At 96 accelerated iterations, warehouse and sortation attain 42.31 and 44.20
tasks per step in \revbegin{}6.25 and 5.80\revend{}~ms. Their throughput, and that on the two random
layouts, differs from the \rev{result of Table~\ref{tab:throughput}} by less than 0.004 tasks per step.
At 288 accelerated iterations, Berlin and Paris reach 33.86 and 34.24 tasks per step.

\begin{figure}[!t]
\centering
\includegraphics[width=\linewidth]{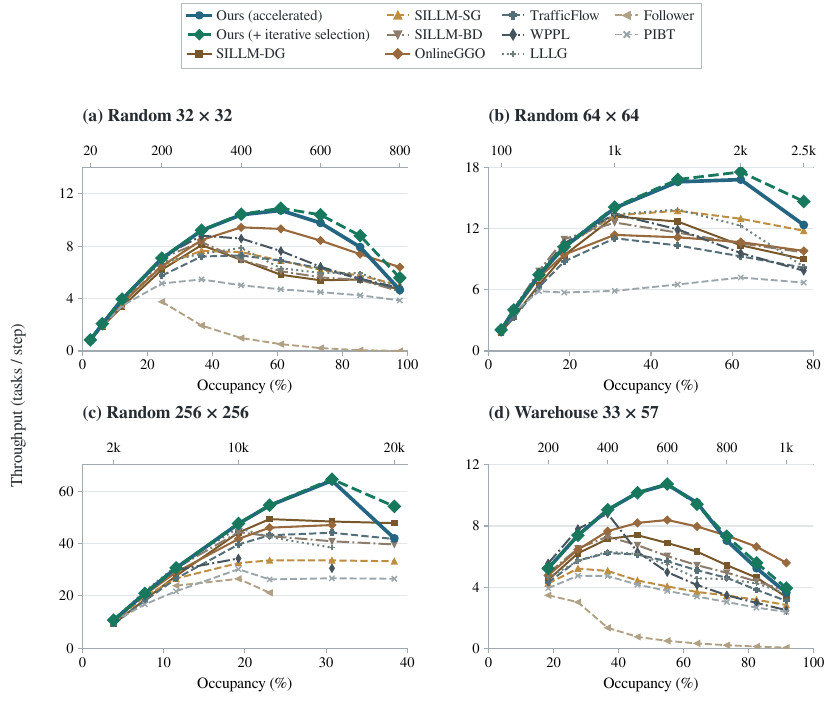}
\caption[Population transfer with accelerated allocation and iterative action selection]{\protect\input{figures/density_appendix_refresh_20260926/selected_caption.tex}}
\label{fig:density-extended}
\end{figure}

\paragraph{Population and iteration count.}
Figure~\ref{fig:density-extended} extends the population study to the warehouse
and \revbegin{}compares both action-selection rules across the full population range.
Table~\ref{tab:density-budgets} reports the throughput--cost tradeoff across
allocation budgets with the same trained policy. At \num{20000} agents,
864 iterations with iterative selection attain 48.04 tasks per step at
61.30~ms per step, compared with SILLM-DG's 47.77 tasks per step at
465~ms per step. Larger budgets provide further
throughput gains, with their costs reported in Table~\ref{tab:density-budgets}.\revend{}
\rev{At \num{4608} iterations with iterative selection, the three training seeds
attain 54.9, 40.0 and 67.5 tasks per step; throughput at this density therefore
varies across training seeds.}
\revbegin{}For the population curves, baseline counts are 288\revend{} on random maps and 144 on
the warehouse; \revbegin{}higher counts follow the offline probability-accuracy calibration
below.\revend{}
Our points average three training seeds and eight cases per seed.

\begin{table}[htbp]
\centering
\small
\revbegin
\begingroup
\setlength{\tabcolsep}{3pt}
\begin{tabular}{lrrrrrrrr}
& & & \multicolumn{2}{c}{Accelerated} & \multicolumn{2}{c}{+ iterative selection} & \multicolumn{2}{c}{OnlineGGO} \\
\cmidrule(lr){4-5}\cmidrule(lr){6-7}\cmidrule(lr){8-9}
Map & Agents & Iterations & Tasks/step & ms/step & Tasks/step & ms/step & Tasks/step & ms/step \\
\midrule
Random $32^2$ & 700 & 288 & 7.67 & 8.30 & 8.50 & 8.67 & 7.39 & 6.10 \\
 & & 576 & 7.86 & 15.34 & 8.79 & 15.71 &  &  \\
\addlinespace[2pt]
Random $32^2$ & 800 & 288 & 3.97 & 8.76 & 4.75 & 9.47 & 6.39 & 6.68 \\
 & & 2,304 & 4.65 & 41.50 & 5.58 & 42.29 &  &  \\
\addlinespace[2pt]
Random $64^2$ & 2,500 & 288 & 11.46 & 9.05 & 14.00 & 9.64 & 9.82 & \textemdash \\
 & & 2,304 & 12.37 & 51.30 & 14.66 & 51.97 &  &  \\
\addlinespace[2pt]
Random $256^2$ & 20,000 & 288 & 32.64 & 23.16 & 41.28 & 24.13 & \textemdash & \textemdash \\
 & & 864 & 36.89 & 61.00 & 48.04 & 61.30 &  &  \\
 & & 4,608 & 41.98 & 304.42 & 54.14 & 300.76 &  &  \\
\addlinespace[2pt]
Warehouse & 700 & 144 & 9.50 & 3.90 & 9.66 & 4.28 & 7.97 & 7.52 \\
 & & 288 & 9.53 & 6.87 & 9.42 & 7.22 &  &  \\
\addlinespace[2pt]
Warehouse & 800 & 144 & 4.49 & 4.44 & 4.89 & 4.73 & 7.37 & 8.28 \\
 & & 4,608 & 6.86 & 82.11 & 7.33 & 82.44 &  &  \\
\addlinespace[2pt]
Warehouse & 900 & 144 & 2.07 & 4.66 & 2.12 & 5.07 & 6.64 & 9.35 \\
 & & 4,608 & 5.23 & 84.07 & 5.59 & 84.63 &  &  \\
\addlinespace[2pt]
Warehouse & 1,000 & 144 & 1.12 & 5.56 & 0.99 & 5.79 & 5.59 & 10.20 \\
 & & 2,304 & 3.51 & 44.10 & 3.94 & 44.89 &  &  \\
\bottomrule
\end{tabular}
\endgroup

\caption[Allocation budgets, action selection, throughput and cost]{\protect Action selection at selected Sinkhorn budgets. Our two action-selection variants use matched checkpoints, cases and allocation updates; throughput averages three training seeds and eight cases per seed. All costs use single-instance measurements (Appendix~\ref{app:cost}). Dashes denote unavailable measurements.
}
\label{tab:density-budgets}
\end{table}

On the $33\times57$ warehouse with 700 agents, 144 accelerated iterations attain \revbegin{}9.50\revend{} tasks per
step, compared with 2.63 at 144 plain iterations and 9.51 at \num{4032} plain
iterations. At 600 agents on $32^2$ and \num{2500} on $64^2$, acceleration at
288 iterations raises throughput from 6.58 to 9.77 and from 7.71 to 11.49
\revbegin{}(Figure~\ref{fig:density}).\revend{}
These comparisons show that changing the allocation updates can improve throughput
under congestion without retraining.

\paragraph{Competition across scales.}
We connect agents through destination cells to which they assign probability
greater than 0.01 and whose total assigned probability exceeds 0.95.
Table~\ref{tab:sinkhorn_competition} reports the largest connected group's size
and approximate graph diameter, measured in agent--cell--agent links.
Training measurements use the saved states and live actors at \num{3000},
\num{10000} and \num{24999} updates, with 128 streams per checkpoint and policy.
Evaluation measurements use 25 evenly spaced states from each of ten
\revbegin{}runs of \num{1000} timesteps,\revend{} with training seed~0 and case~0.
The largest group averages about four agents during training, compared with
\num{3641} at \num{20000} agents on $256^2$.
On the 900-agent warehouse, its mean spatial span is 49 by 26 cells on the
$33\times57$ map. Increasing population can therefore connect local competition
over distances substantially larger than those observed during training.

\begin{table}[htbp]
\centering
\small
\begin{tabular}{lrrr}
Setting & Group size & Diameter & $K^\star$ \\
\midrule
Unstructured training & 4.34 & 2.25 & --- \\
Structured training & 4.03 & 2.51 & --- \\
\midrule
Warehouse, 600 & 13.84 & 6.36 & 144 \\
Warehouse, 700 & 49.68 & 17.56 & 288 \\
Warehouse, 800 & 314.68 & 53.52 & \num{4608} \\
Warehouse, 900 & 456.36 & 60.68 & \num{4608} \\
Warehouse, \num{1000} & 675.80 & 68.68 & \num{2304} \\
Random $32^2$, 700 & 96.52 & 24.00 & 576 \\
Random $32^2$, 800 & 521.28 & 56.96 & \num{2304} \\
Random $64^2$, \num{2500} & 816.60 & 60.36 & \num{2304} \\
Random $256^2$, \num{16000} & 32.76 & 12.72 & 288 \\
Random $256^2$, \num{20000} & \num{3641.16} & 91.00 & \num{4608} \\
\bottomrule
\end{tabular}

\caption{Connected competition in historical training and evaluation states.
Group sizes and graph diameters are means over sampled states, using an
\num{18432}-iteration allocation reference. $K^\star$ is the smallest eligible
count selected by the empirical allocation-error criterion; training rows report no
evaluation count. Diameter is estimated by two successive breadth-first searches.}
\label{tab:sinkhorn_competition}
\end{table}

\paragraph{Offline iteration calibration.}
\revbegin{}For the population curves, we calibrate iteration counts on saved states by\revend{}
comparing allocation probabilities with those from a much longer run.
\revbegin{}Each selected budget remains fixed over all timesteps.\revend{}
Holding each state's inputs fixed, let \revbegin{}$\pi_{i,K}(\cdot\mid s)$\revend{} be agent
$i$'s allocation distribution after $K$ iterations per pass. We measure
\begin{equation}
 e_K = \frac{1}{|\mathcal S|}\sum_{s\in\mathcal S}
 \revbegin{}\frac{1}{2N(s)}\sum_{i=1}^{N(s)}
 \left\|\pi_{i,K}(\cdot\mid s)-\pi_{i,K_{\rm ref}}(\cdot\mid s)\right\|_1,\revend{}
 \label{eq:allocation_accuracy}
\end{equation}
where $\mathcal S$ is the set of sampled states, \revbegin{}$N(s)$\revend{} is the number of active
agents, and $K_{\rm ref}=\num{18432}$.
Thus $e_K$ measures the average difference in allocation probabilities over agents
and sampled states. We use 0.005 as an empirical tolerance for this comparison;
the longer run is a finite numerical reference.
Among counts with complete three-seed, eight-case evaluations, we choose the
smallest at or above the family baseline for which $e_K\leq 0.005$ and remains
within this tolerance at larger tested counts. For example, on $256^2$ we select 288 iterations at \num{16000} agents
and \num{4608} at \num{20000} agents.
The \num{9216}- and \num{18432}-iteration references differ by less than
0.0005 in mean total variation in every evaluated condition.

Training occupancy spans 16.8--34.2\% for the unstructured pool and
21.7--68.4\% for the structured pool; the highest evaluated occupancies
also extend beyond these ranges.
\revbegin{}The larger connected groups provide a possible explanation for the increased
allocation budgets at high occupancy.\revend{}

\FloatBarrier
\subsection{Long-horizon evaluation}
\label{app:long_horizon}

\begin{figure}[!t]
\centering
\includegraphics[width=\linewidth]{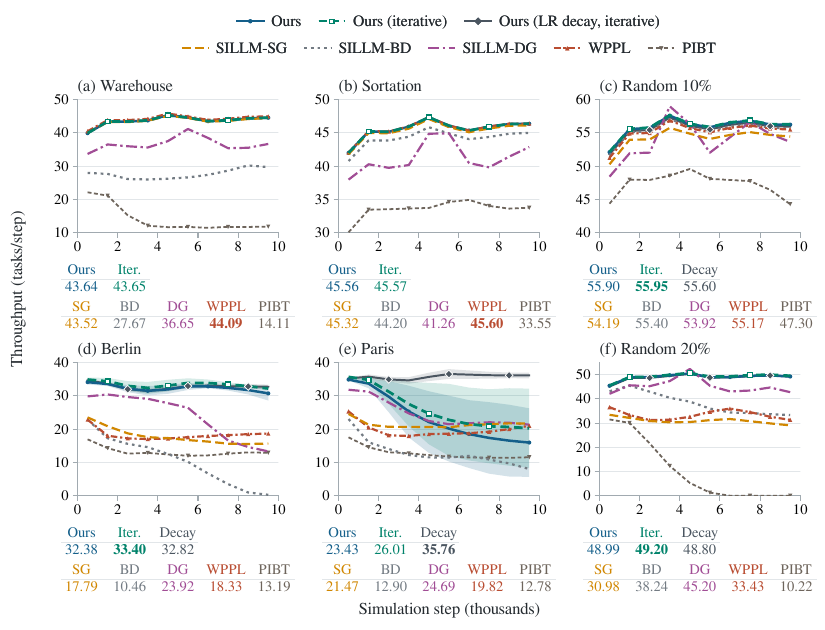}
\revbegin
\caption[Continuation throughput with SILLM and WPPL on six layouts]{\protect\input{figures/six_map_sillm_20261008/caption.tex}}
\label{fig:long-horizon-six-maps}
\end{figure}

\revbegin{}We extend evaluation to\revend{} \num{10000} uninterrupted steps with \num{10000} agents
\revbegin{}to examine whether throughput persists under the transfer protocol of
Table~\ref{tab:throughput}.
For each of three training seeds, the same terminal checkpoint is reused across
all layouts within its class, without target-layout retraining or checkpoint selection.
Both action-selection rules use 144 accelerated Sinkhorn iterations per pass;
learned parameters remain fixed.
WPPL retains the \rev{setting of Table~\ref{tab:throughput}}, \num{40000} LNS iterations, eight threads
and a 15-step planning window.
Standard-horizon execution costs are reported in
Tables~\ref{tab:throughput} and~\ref{tab:sinkhorn_deployment}.
Figure~\ref{fig:long-horizon-six-maps} compares all six layouts with the three
SILLM variants, WPPL and PIBT. Our policy closely tracks SILLM-SG and WPPL on warehouse
and sortation; on Berlin and both random layouts, its whole-run mean exceeds
those of all five baselines.
Iterative selection raises mean throughput from 32.38 to 33.40 on Berlin and
from 23.43 to 26.01 on Paris, where SILLM-DG has the highest baseline mean of 24.69.\revend{}

\revbegin
\paragraph{Decline on Berlin and Paris.}
Depending on the training seed, our policy either sustains its throughput on the
two city layouts or builds up congestion that it does not resolve, especially on
Paris (Figure~\ref{fig:long-horizon-six-maps}). We observed this congestion in the
right part of the Paris map and in the lower-left corner of Berlin, regions that
we interpret as large cul-de-sacs, which traffic must leave the way it entered.
Iterative action selection raises mean throughput on both layouts but does not
eliminate the congestion. Policies retrained with the learning rate decayed to one
tenth over the last \num{10000} updates, deployed with their final weights and
iterative selection, sustain their throughput in every case for all three seeds
on both layouts (Figure~\ref{fig:long-horizon-six-maps}d,e). This suggests that
there is further room to improve long-horizon stability through optimisation,
which we leave to future work.
\revend

\FloatBarrier
\subsection{Additional large layouts}
\label{app:additional_layouts}

Table~\ref{tab:additional_layouts} evaluates the same unstructured policy on two
$256^2$ random layouts and the $194^2$ Dragon Age: Origins map \texttt{ost003d}
from the Moving AI benchmark~\citep{sturtevant2012benchmarks}.
Random 30\% retains the largest connected component of a 30\%-obstacle draw;
Random 40\% uses clustered obstacles. The three maps contain \num{45049},
\num{39322} and \num{13214} free cells, respectively.

Each method uses the same eight cases per layout and \num{2500} scored steps.
Our policy uses terminal averaged weights from three training seeds and 288
plain allocation iterations. \rev{On all three layouts,} SILLM-BD and SILLM-DG
transfer their released \revbegin{}IL\revend{} checkpoints trained on 20\%-obstacle random maps,
TrafficFlow uses GP-R100-Re10-F2, and LLLG uses its \num{10000}-agent setting
(planning window 8, horizon 5, two refinements, 10-second limit).
WPPL uses \num{2500} LNS iterations on these layouts\rev{; on \texttt{ost003d} it runs without
map weights, as SILLM's release provides none for this map}.
PIBT uses SILLM's implementation \rev{and hm+GPIBT uses OnlineGGO's handcrafted guidance on every layout.
For hm+GPIBT on Random 30\% we raised the planning time limit per step of its released driver
from 10 to 20 seconds.} All learned policies retain their trained
parameters.

\begin{table}[htbp]
\centering
\scriptsize
\setlength{\tabcolsep}{2pt}
\begin{tabular}{@{}lrrrrrrrrr@{}}
\toprule
Layout & Agents & Ours & SILLM-BD & SILLM-DG & LLLG & TrafficFlow & PIBT & WPPL & hm+GPIBT \\
\midrule
Random 30\% (LCC) & \num{10000} & $18.86\!\pm\!2.26$ & \rev{12.83} & \rev{18.62} & 16.68 & 17.50 & 13.91 & 14.26 & \rev{15.15} \\
Random 40\% (clustered) & \num{10000} & $22.10\!\pm\!1.02$ & \rev{9.64} & \rev{14.49} & 12.05 & 13.50 & 11.88 & 14.33 & \rev{14.20} \\
\texttt{ost003d} & \num{6000} & $6.09\!\pm\!0.53$ & \rev{4.95} & \rev{4.29} & \rev{4.43} & \rev{4.00} & 3.78 & \rev{4.02} & 3.90 \\
\bottomrule
\end{tabular}

\caption{Throughput on additional large layouts, in tasks per step. Our entries
show mean $\pm$ sample standard deviation across three training seeds; each seed
and each baseline average eight matching cases.}
\label{tab:additional_layouts}
\end{table}

\revbegin{}The same unstructured policy transfers to these more obstructed layouts without
retraining, attaining the highest mean throughput among the evaluated methods
on all three maps.\revend{}

\revbegin
\paragraph{Released RAILGUN checkpoint.}
RAILGUN reports experiments with up to 256 agents~\citep{tang2025railgun}. Our harness
reproduces the results distributed with its release on its own warehouse benchmark
(individual success rate 0.989 against 0.990 at 32 agents). On the 128 small random maps of
Table~\ref{tab:small_all} with 8 agents, using the start and goal files of the release,
its checkpoint completes 0.52 tasks per step in lifelong operation, against 0.61 for our
policy. As a transfer check far outside its reported range, we also ran it on the six
\num{10000}-agent layouts with eight cases each, where mean throughput is between 0.001
and 0.010 tasks per step. Its observation encodes agents by integer index and goals by
unnormalised shortest-path cost, both of which exceed their training range at this
scale. Rescaling these two channels to the training range gives 1.2--5.6 tasks per
step, which suggests that input scale accounts for part of the failure.
\revend

\FloatBarrier
\revbegin
\subsection{Training pools across layout classes}
\label{app:training_pools}

Table~\ref{tab:training_pools} evaluates each class-specific policy on the layouts
of the other class, and a single policy trained on both pools, under the protocol
of Table~\ref{tab:throughput}. The combined pool contains 48 structured and 80
unstructured maps and keeps the pool size and training budget of one
class-specific run. A policy trained on the structured pool fails on Berlin and
Paris, and one trained on the unstructured pool is weaker on the warehouse.
A single policy trained on both pools approaches the class-specific policies on
four layouts and remains weaker on Berlin and Paris. We leave a training recipe
that closes this gap to future work.

\begin{table}[htbp]
\centering
\small
\revbegin
\begin{tabular}{@{}lrrrrrr@{}}
\toprule
Training pool & Warehouse & Sortation & Berlin & Paris & Random 10\% & Random 20\% \\
\midrule
Structured & 42.31 & 44.20 & 7.22 & 10.68 & 54.24 & 46.04 \\
Unstructured & 38.75 & 43.40 & 32.98 & 32.97 & 54.19 & 47.52 \\
Both & 41.64 & 44.05 & 31.77 & 29.19 & 54.23 & 47.36 \\
\bottomrule
\end{tabular}

\caption{Throughput at \num{10000} agents by training pool, in tasks per step.
The structured and unstructured rows average three training seeds with eight
cases each; the policy trained on both pools has one training seed. Entries for a
pool's own class are repeated evaluations and agree with Table~\ref{tab:throughput}
within 0.2 tasks per step.}
\label{tab:training_pools}
\end{table}
\revend

\FloatBarrier
\section{Mechanism controls and supporting evidence}
\label{app:mechanism_controls}

The following controls use separate runs from the structured and unstructured
training pools in Appendix~\ref{app:training}; \revbegin{}the reference-only control in
Table~\ref{tab:field_controls} reuses the trained policies.\revend{} Deployment evaluations use
terminal averaged weights; learning curves use current parameters.

\subsection{Actor structure at deployment scale}
\label{app:actor_ablations}

Table~\ref{tab:actor_ablations} compares the full actor with six retrained variants
on the six deployment layouts at \num{10000} agents. Each variant retains its
policy family's training pool and \num{25000}-update budget. We evaluate terminal
averaged weights from three training seeds on eight cases per layout, using
\num{3200} steps and 144 plain allocation iterations on structured maps, and
\num{2500} steps and 288 iterations on unstructured maps.

\paragraph{Controls.}
The depth variants use $H=1$ or $H=2$ during both training and evaluation.
Removing allocation replaces both Sinkhorn passes and the reverse-move penalty
with independent normalisation of route affinities, leaving the simulator unchanged.
The combined variant also sets $H=1$.
Removing the vector branch sets $G=0$ in Eq.~\eqref{eq:routegain}, retaining
the action-score branch, goal-conditioned gains and depth corrections.
The no-gain control fixes $\gamma_h(x)=0$ in Eq.~\eqref{eq:read}, retaining
the shared scores, vector branch, depth corrections, geometric reference and allocation.

\begin{table}[htbp]
\centering
\footnotesize
\setlength{\tabcolsep}{2pt}
\begin{tabular}{@{}lrrrrrr@{}}
\toprule
& \multicolumn{2}{c}{Structured} & \multicolumn{4}{c}{Unstructured} \\
\cmidrule(lr){2-3}\cmidrule(l){4-7}
Actor & Warehouse & Sortation & Berlin & Paris & Random 10\% & Random 20\% \\
\midrule
Full ($H=4$) & $42.31\!\pm\!0.06$ & $44.20\!\pm\!0.02$ & $32.99\!\pm\!0.83$ & $33.17\!\pm\!1.90$ & $54.19\!\pm\!0.04$ & $47.53\!\pm\!0.04$ \\
$H=1$ & $40.06\!\pm\!1.60$ & $43.90\!\pm\!0.04$ & $30.23\!\pm\!1.37$ & $33.33\!\pm\!1.89$ & $53.80\!\pm\!0.01$ & $45.99\!\pm\!0.04$ \\
$H=2$ & $41.64\!\pm\!0.05$ & $44.06\!\pm\!0.03$ & $26.73\!\pm\!3.00$ & $32.13\!\pm\!2.59$ & $54.14\!\pm\!0.05$ & $46.95\!\pm\!0.06$ \\
Without allocation & $40.77\!\pm\!2.26$ & $44.10\!\pm\!0.07$ & $18.76\!\pm\!3.44$ & $22.66\!\pm\!3.41$ & $53.40\!\pm\!0.04$ & $44.81\!\pm\!0.34$ \\
$H=1$, no allocation & $22.24\!\pm\!16.61$ & $43.82\!\pm\!0.05$ & $20.65\!\pm\!0.61$ & $20.79\!\pm\!1.69$ & $53.04\!\pm\!0.05$ & $42.91\!\pm\!0.33$ \\
Without vector field $G$ & $42.18\!\pm\!0.09$ & $44.20\!\pm\!0.01$ & $29.33\!\pm\!3.34$ & $31.24\!\pm\!2.82$ & $54.27\!\pm\!0.03$ & $47.66\!\pm\!0.05$ \\
Without gains $\gamma$ & $42.00\!\pm\!0.17$ & $44.23\!\pm\!0.03$ & $29.17\!\pm\!1.91$ & $31.29\!\pm\!1.81$ & $54.34\!\pm\!0.05$ & $47.91\!\pm\!0.04$ \\
\bottomrule
\end{tabular}

\caption{Actor-structure ablations at \num{10000} agents. Entries are throughput
in tasks per step, reported as mean $\pm$ sample standard deviation over three
training seeds; each seed averages eight cases.}
\label{tab:actor_ablations}
\end{table}

Four departures improve mean throughput over one on warehouse, sortation and
both random maps; the city results do not follow a monotonic depth ordering.
Removing allocation reduces Berlin throughput from 32.99 to 18.76 and Paris
from 33.17 to 22.66. On warehouse, combining $H=1$ with allocation removal gives a larger mean
throughput reduction than either change alone. Removing the vector branch leaves structured and random-map throughput
close to the full actor, while reducing throughput on both cities.
Appendices~\ref{app:continuation_control} and~\ref{app:allocation_control}
examine continuation depth and allocation across populations.

\paragraph{Goal-conditioned gains.}
Removing the gains reduces mean throughput from 42.31 to 42.00 on warehouse,
from 32.99 to 29.17 on Berlin, and from 33.17 to 31.29 on Paris.
Sortation throughput is nearly unchanged, while both random-map means increase
slightly. The benefit of these gains therefore depends on the layout.

\revbegin
\paragraph{Controls for the learned field.}
Table~\ref{tab:field_controls} adds three controls for the learned field and the
route operator, under the protocol of Table~\ref{tab:throughput}. The
reference-only control keeps the soft shortest-path reference, the allocation
and the learned temperatures of the trained policies and removes the field; no
parameter is retrained. Throughput then falls to 0.3--4.5 tasks per step, and on
the warehouse 92\% of agents complete no task. The direct-score control is
retrained with the same inputs, and its five action scores enter the allocation
directly, without the reference or the route operator. It is 1.4--5.5 tasks per
step below the full actor and 3--9\% below the highest-throughput baseline of
Table~\ref{tab:throughput} on every layout. One-step reweighting, also retrained,
uses $H=1$ without goal-conditioned gains or the vector branch, so the scores
reweight the reference for the next move only. It recovers most of this gap on
sortation, Paris and the random maps; on the warehouse and Berlin it remains 5.3
and 4.1 tasks per step below the full actor and varies across training seeds.

\begin{table}[htbp]
\centering
\footnotesize
\setlength{\tabcolsep}{2pt}
\revbegin
\begin{tabular}{@{}lrrrrrr@{}}
\toprule
& \multicolumn{2}{c}{Structured} & \multicolumn{4}{c}{Unstructured} \\
\cmidrule(lr){2-3}\cmidrule(l){4-7}
Variant & Warehouse & Sortation & Berlin & Paris & Random 10\% & Random 20\% \\
\midrule
Reference only & $0.28\!\pm\!0.00$ & $2.88\!\pm\!0.05$ & $0.80\!\pm\!0.01$ & $0.96\!\pm\!0.01$ & $4.50\!\pm\!0.31$ & $0.39\!\pm\!0.02$ \\
Direct scores & $39.07\!\pm\!0.59$ & $42.82\!\pm\!0.15$ & $27.46\!\pm\!0.56$ & $28.71\!\pm\!0.62$ & $51.73\!\pm\!0.13$ & $43.56\!\pm\!0.12$ \\
One-step reweighting & $37.05\!\pm\!3.45$ & $43.82\!\pm\!0.02$ & $28.94\!\pm\!7.19$ & $32.92\!\pm\!1.90$ & $53.64\!\pm\!0.04$ & $45.82\!\pm\!0.10$ \\
Full ($H=4$) & $42.31\!\pm\!0.06$ & $44.20\!\pm\!0.02$ & $32.99\!\pm\!0.83$ & $33.17\!\pm\!1.90$ & $54.19\!\pm\!0.03$ & $47.52\!\pm\!0.04$ \\
\bottomrule
\end{tabular}

\caption{Controls for the learned field at \num{10000} agents, under the
protocol of Table~\ref{tab:throughput}. Entries are throughput in tasks per
step, mean $\pm$ sample standard deviation over three training seeds; each seed
averages eight cases. The full row is that of Table~\ref{tab:throughput}, with 144
allocation iterations on the random layouts where Table~\ref{tab:actor_ablations}
uses 288.}
\label{tab:field_controls}
\end{table}
\revend

\subsection{Learning progress}
\label{app:actor_learning}

\begin{figure}[htbp]
\centering
\includegraphics[width=\linewidth]{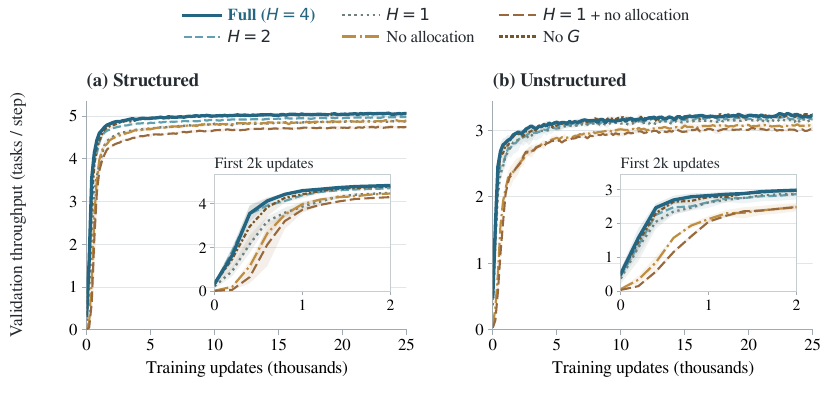}
\caption[Validation throughput during training]{\protect\input{figures/actor_training_progress_20260925/caption.tex}}
\label{fig:actor_learning}
\end{figure}

Figure~\ref{fig:actor_learning} tracks validation throughput for selected variants in
Table~\ref{tab:actor_ablations}. Current parameters are evaluated every 200 updates
on 32 procedural validation maps generated separately from the training pool,
using 64 reset cases, deterministic proposals and 128 steps per case, with the
first 15 excluded from scoring. Each training update collects $128\times64$
joint environment transitions.

On structured validation, $H=4$ outperforms $H=1$ from early training onward.
Allocation improves early validation in both families, with a larger
separation on unstructured maps that narrows as training progresses. Removing
$G$ has little effect on late validation in either family, despite the lower
city throughput in Table~\ref{tab:actor_ablations}.

\FloatBarrier

\subsection{Continuation depth}
\label{app:continuation_control}

\paragraph{Training controls.}
The full policies use $H=4$ departures. Depth controls set $H=1$ or $H=2$ during
both training and evaluation, retaining their respective fixed training pools
and the other training settings in Appendix~\ref{app:training}. Each run completes
\num{25000} training iterations.
The convolutional network still observes the same traffic inputs at every depth.

\paragraph{Learning and deployment.}
Figure~\ref{fig:mechanisms}a uses three independent structured-policy training runs
per depth, \revbegin{}following the validation protocol in Appendix~\ref{app:actor_learning}
with 48 plain allocation iterations per pass.\revend{} Curves show equal-run means and
sample standard deviations without smoothing; \revbegin{}gaps indicate incomplete seed
coverage. The common training batch makes update counts comparable across depths.\revend{}

The annotations use terminal averaged weights,
evaluated on eight large-warehouse instances per training run with \num{10000}
agents, \num{3200} steps, 144 allocation iterations and layered reference tables.
All steps contribute to throughput. Cases are averaged within each run before
computing the mean and sample standard deviation across the three runs.

\paragraph{Corridor traffic.}
Figure~\ref{fig:mechanisms}b uses terminal averaged policies from three independent
training runs per depth, evaluated separately on eight $33\times57$ warehouse
instances with 400 agents and \num{3200} scored steps. Execution uses deterministic
proposals, 144 allocation iterations and layered reference tables. We identify
all 172 narrow passages from the walls alone: 84 horizontal passages with three
interior cells and 88 vertical passages with two.

For neighbouring occupied cells, let $q_i(j)$ be agent $i$'s probability of moving
into agent $j$'s cell after route reweighting and before allocation. Opposing
demand is the mean of $q_i(j)q_j(i)$ over occupied internal passage edges and time
steps. We first compute this rate within each case, then average equally over
the eight cases and three training runs. It measures incompatible route preferences, not executed collisions; outward movement by both agents is valid.
The traffic illustration shows actual moves and waits in passage 122, training
run 0, case 0, at steps 1024--1151. The central passage and time window were fixed
before inspecting the results. Colours track the entry side, not the current
direction. Diamonds mark goal completions because passage cells can be goals,
and returning towards the entry can therefore be productive.

At 400 agents, opposing demand falls by 39.4\% with $H=4$ and throughput rises
by 9.4\%. Actual axial moves per occupied passage-cell step remain similar
(87.97\% and 87.59\% for $H=1,4$). A lower-load control at 200 agents gives smaller
improvements of 5.9\% in opposing demand and 2.9\% in throughput, with overlapping
training-run ranges for demand.

To distinguish the visited traffic states from the immediate effect of the route operator,
we also apply all six intact policies to the same \num{1152} states, sampled
every 128 steps from 128 to \num{3072} across all 48 policy--case \revbegin{}pairs.\revend{}
On these common states, mean opposing demand is 5.69\% for $H=1$ and 5.36\%
for $H=4$, with overlapping training-run ranges. Both depth variants produce lower
opposing demand on states generated by $H=4$. The larger difference \revbegin{}above\revend{}
concerns the traffic produced over time; the immediate difference on
identical states is smaller.

\paragraph{Evaluation under load.}
The following population comparisons use terminal averaged weights.
The structured policy is evaluated on the $33\times57$ warehouse with alternating
workstation--storage tasks; the unstructured policy uses the $32\times32$ random
map with uniform tasks. The maps contain \num{1091} and \num{819} free cells,
respectively. Occupancy is population divided by this count. Each point uses
eight cases \revbegin{}of \num{1000} timesteps and deterministic proposals; all timesteps\revend{} count.
Plain allocation uses 144 iterations on warehouse and 288 on random, fixed across
populations and continuation depths. Reference tables use the layered and sweep builders,
respectively. Within each map, full and depth-control evaluations use the same
initial-state and task files.

Table~\ref{tab:continuation_depth} reports throughput by training seed.
The comparisons below average training seeds 0--2.

At approximately 37\% occupancy, $H=4$ raises mean throughput over $H=1$ by
\revbegin{}9.2\% on warehouse and 8.5\%\revend{} on random. The depths give similar throughput at
light load on random; depth ordering becomes nonmonotonic at the highest loads,
where warehouse throughput also deteriorates.

\begin{table}[htbp]
\centering
\small
\setlength{\tabcolsep}{4pt}
\begingroup
\scriptsize
\setlength{\tabcolsep}{3pt}
\begin{tabular}{lrr*{9}{r}}
\toprule
 & & & \multicolumn{3}{c}{$H=1$} & \multicolumn{3}{c}{$H=2$} & \multicolumn{3}{c}{$H=4$} \\
Map & Agents & Occ. (\%) & s0 & s1 & s2 & s0 & s1 & s2 & s0 & s1 & s2 \\
\midrule
Warehouse & 200 & 18.3 & 5.092 & 5.100 & 5.086 & 5.141 & 5.158 & 5.129 & 5.223 & 5.246 & 5.225 \\
 & 300 & 27.5 & 6.943 & 6.991 & 6.973 & 7.070 & 7.145 & 7.055 & 7.382 & 7.440 & 7.340 \\
 & 400 & 36.7 & 8.163 & 8.364 & 8.329 & 8.440 & 8.654 & 8.426 & 9.052 & 9.103 & 8.984 \\
 & 500 & 45.8 & 8.667 & 9.192 & 7.203 & 6.533 & 9.676 & 9.165 & 9.925 & 9.470 & 10.069 \\
 & 600 & 55.0 & 3.509 & 3.574 & 2.089 & 3.086 & 6.167 & 5.055 & 5.030 & 3.266 & 9.561 \\
\midrule
Random & 20 & 2.4 & 0.855 & 0.853 & 0.847 & 0.849 & 0.853 & 0.849 & 0.844 & 0.835 & 0.836 \\
 & 50 & 6.1 & 2.090 & 2.088 & 2.088 & 2.094 & 2.092 & 2.084 & 2.063 & 2.056 & 2.061 \\
 & 100 & 12.2 & 3.931 & 3.944 & 3.940 & 3.967 & 3.975 & 3.953 & 3.943 & 3.949 & 3.933 \\
 & 200 & 24.4 & 6.755 & 6.830 & 6.805 & 6.947 & 6.997 & 6.959 & 7.042 & 7.051 & 7.072 \\
 & 300 & 36.6 & 8.311 & 8.569 & 8.526 & 8.865 & 8.902 & 8.919 & 9.225 & 9.050 & 9.302 \\
 & 500 & 61.1 & 8.761 & 9.338 & 8.965 & 7.694 & 7.177 & 9.852 & 10.622 & 10.474 & 10.861 \\
\bottomrule
\end{tabular}
\endgroup

\caption{Throughput by continuation depth and training seed. Each entry averages
the same eight cases for its layout and population.}
\label{tab:continuation_depth}
\end{table}

\subsection{Destination allocation under load}
\label{app:allocation_control}

The allocation control replaces both allocation passes and the reverse-move penalty
with independent normalisation of next-move affinities at the learned allocation
temperature. The convolutional network, goal-conditioned route calculation and simulator
are unchanged. Controls use the unstructured training pool, with the full policy's
batch, rollout length, update count and optimiser settings.

Table~\ref{tab:allocation_current} compares terminal averaged weights at
populations \num{6000}, \num{10000} and \num{12000}, on the two $256^2$ random
maps. Each cell averages eight released start/task cases over \num{2500} steps,
with argmax proposals, layered route tables and 288 plain allocation iterations in
the full policy. Results average three training runs per condition (seeds 0--2).
\revbegin{}The policies have similar throughput at \num{6000} agents. At \num{12000} agents
on Random 20\%, allocation preserves throughput as demand increases: the full
policy attains 54.47 tasks per step, compared with 27.85 without allocation.\revend{}

\begin{table}[htbp]
\centering
\small
\setlength{\tabcolsep}{4pt}
\begin{tabular}{lrrrrrrr}
\toprule
& & \multicolumn{3}{c}{Full} & \multicolumn{3}{c}{Without allocation} \\
\cmidrule(lr){3-5}\cmidrule(lr){6-8}
Layout & Agents & Seed 0 & Seed 1 & Seed 2 & Seed 0 & Seed 1 & Seed 2 \\
\midrule
Random 20\% & 6,000 & 30.63 & 30.67 & 30.62 & 30.20 & 30.28 & 30.25 \\
Random 20\% & 10,000 & 47.50 & 47.58 & 47.51 & 44.42 & 45.06 & 44.95 \\
Random 20\% & 12,000 & 54.40 & 54.52 & 54.49 & 22.55 & 32.28 & 28.72 \\
\midrule
Random 10\% & 6,000 & 33.25 & 33.26 & 33.26 & 33.14 & 33.17 & 33.16 \\
Random 10\% & 10,000 & 54.15 & 54.23 & 54.19 & 53.35 & 53.44 & 53.40 \\
Random 10\% & 12,000 & 64.03 & 64.14 & 64.09 & 62.43 & 62.60 & 62.53 \\
\bottomrule
\end{tabular}

\caption{Full and allocation-removed policies on sparse and dense random layouts.
Throughput averages eight cases per training seed, over all \num{2500} steps.}
\label{tab:allocation_current}
\end{table}

\end{document}